\documentclass[times,twocolumn,final,authoryear]{elsarticle}

\usepackage{prletters}
\usepackage{framed,multirow}

\usepackage{amssymb}
\usepackage{latexsym}

\usepackage{times}
\usepackage{epsfig}
\usepackage{graphicx}
\usepackage{amsmath}
\usepackage{amssymb}
\usepackage{flushend}
\usepackage{framed,multirow}
\usepackage{amssymb}
\usepackage{amsmath}
\usepackage{latexsym}
\usepackage{mathtools}
\usepackage{algorithm} 
\usepackage{algpseudocode} 
\usepackage{graphicx, caption, subcaption}
\usepackage{amsmath,amssymb}
\usepackage{amsmath}
\usepackage{url}
\usepackage{xcolor}
\definecolor{newcolor}{rgb}{.8,.349,.1}
\newcommand{\R}{\rm I\!R}

\usepackage{setspace}
\usepackage{graphicx}
\usepackage{caption}

\usepackage{soul}

\usepackage{url}
\usepackage{xcolor}
\definecolor{newcolor}{rgb}{.8,.349,.1}

\journal{}

\begin{document}

\begin{frontmatter}

\title{Beyond Neighbourhood-Preserving Transformations for Quantization-Based Unsupervised Hashing}

\author[1]{Sobhan   \snm{Hemati}}
\author[2]{H.R. \snm{Tizhoosh}\corref{cor1}} 
\cortext[cor1]{Corresponding author:}
\ead{tizhoosh@uwaterloo.ca}

\address[1]{Kimia Lab,  University of Waterloo, Waterloo, ON, Canada}
\address[2]{Vector Institute, MaRS Centre, Toronto, ON,  Canada}

\received{}
\finalform{}
\accepted{}
\availableonline{}
\communicated{}

\begin{abstract}
An effective unsupervised hashing algorithm leads to compact binary codes preserving the neighborhood structure of data as much as possible. One of the most established schemes for unsupervised hashing is to reduce the dimensionality of data and then find a rigid (neighborhood-preserving) transformation that reduces the quantization error. Although employing rigid transformations is effective, we may not reduce quantization loss to the ultimate limits. As well, reducing dimensionality and quantization loss in two separate steps seems to be sub-optimal. Motivated by these shortcomings, we propose to employ both rigid and non-rigid transformations to reduce quantization error and dimensionality simultaneously. We relax the orthogonality constraint on the projection in a PCA-formulation and regularize this by a quantization term. We show that both the non-rigid projection matrix and rotation matrix contribute towards minimizing quantization loss but in different ways. A scalable nested coordinate descent approach is proposed to optimize this mixed-integer optimization problem. We evaluate the proposed method on five public benchmark datasets providing almost half a million images. Comparative results indicate that the proposed method mostly outperforms state-of-art linear methods and competes with end-to-end deep solutions.
\end{abstract}


\end{frontmatter}


\section{Introduction}
Search in big image data is one of the major challenges that has many applications in retrieval, classification and clustering of digital images. A well-established approach toward indexing and searching in mass repositories is hashing that assigns binary codes to data records for fast and efficient access. Hashing has been applied to different problems where fast nearest neighbour search is necessary e.g, large-scale clustering \cite{7298596}.

\begin{figure}[htb]
  \centering

  \medskip
    
  \begin{subfigure}[t]{.4\columnwidth}
    \centering\includegraphics[width=\linewidth]{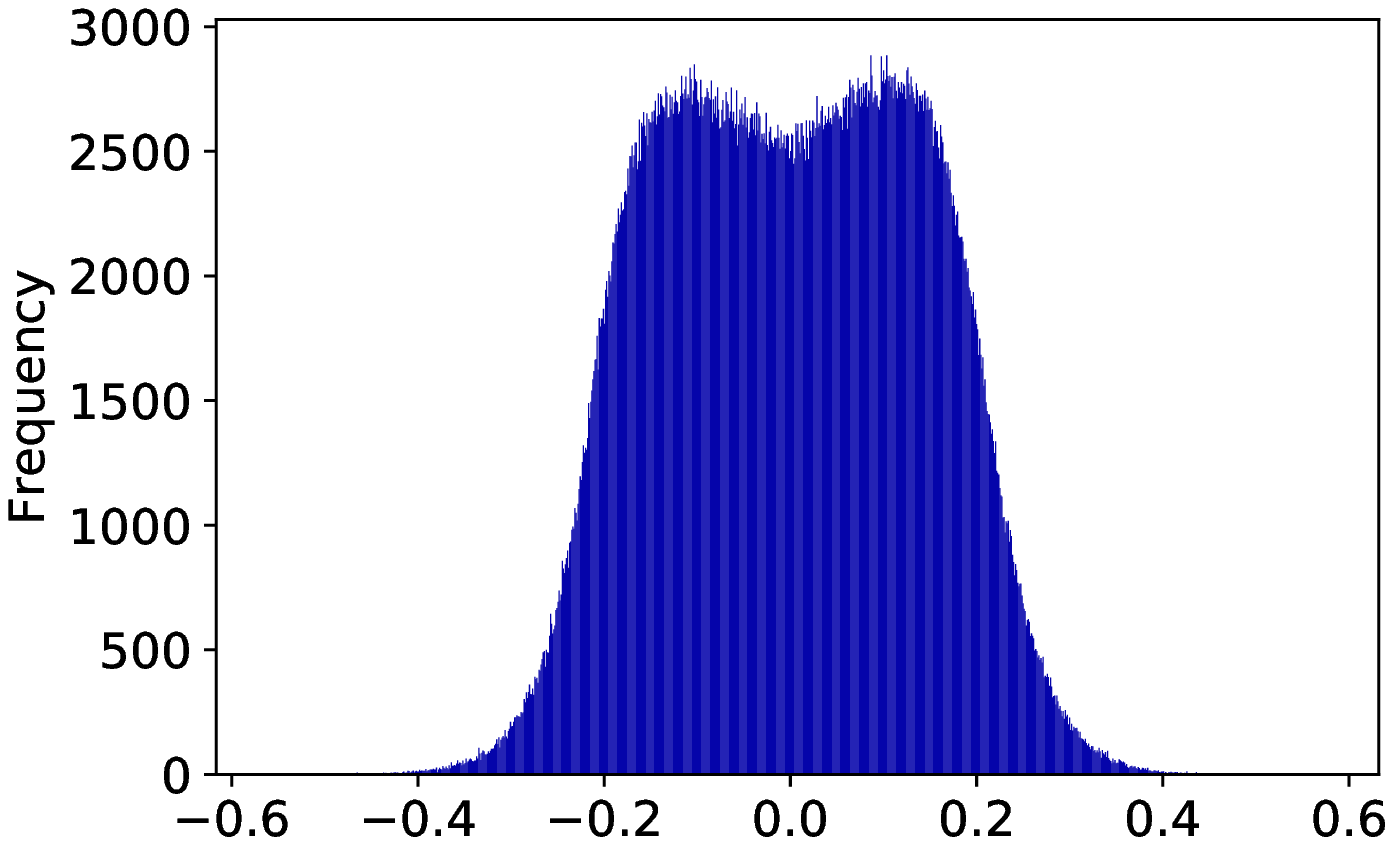}
    \caption{ITQ}
  \end{subfigure}
  \begin{subfigure}[t]{.4\columnwidth}
    \centering\includegraphics[width=\linewidth]{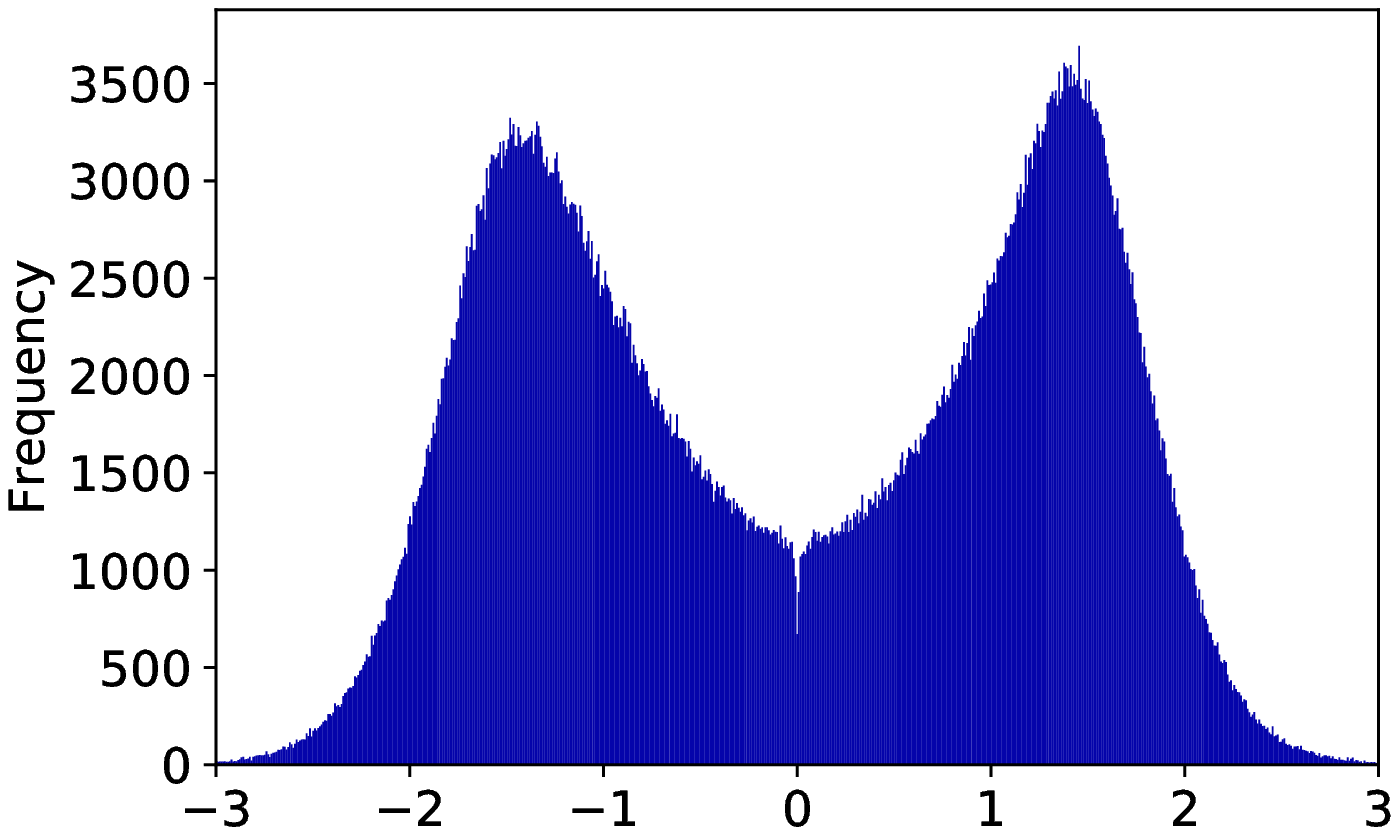}
    \caption{SNRQ}
  \end{subfigure} 
  \caption{Histogram of transformed MNIST dataset using a) ITQ (rigid transformation) and  b) SNRQ (rigid and non-rigid transformations).}%
  \label{fig:1}%
\end{figure}

Hashing algorithms can be categorized in two main sub-groups: data-independent and data-dependent algorithms \cite{wang2017survey}. The locality sensitive hashing (LSH) \cite{4031381} is one of the first representatives in the first group where hash functions are constructed using random projections. Although these methods are faster compared with data-dependent algorithms, their performance is inferior for compact binary codes \cite{pauleve2010locality}. To address this shortcoming, many data-dependent (or learning-based) hashing algorithms which are divided into supervised and unsupervised methods have been proposed.  Supervised hashing methods incorporate label information to learn similarity-preserving binary codes  while in unsupervised hashing, which is the focus of this paper the affinity information in the data is used for code learning \cite{weiss2009spectral}. 

The main challenge in learning similarity-preserving binary codes is the discrete nature of the optimization problems. To tackle this problem different approaches have been proposed including minimizing quantization loss in Iterative Quantization (ITQ)  \cite{gong2012iterative}, Angular Quantization (AQ) \cite{gong2012angular}, equalizing variance across projections in isotropic hashing  (IsoHash) \cite{kong2012isotropic}, reconstruction loss in an encoder-decoder architecture like in binary autoencoder (BA) \cite{carreira2015hashing}, and direct binary code learning in spectral hashing (SH) \cite{weiss2009spectral} and its sparse version \cite{shao2012sparse}. and more recent developments including Anchor Graph Hashing (AGH) \cite{liu2011hashing}, and Discrete Graph Hashing (DGH) \cite{liu2014discrete} Large Graph Hashing with Spectral Rotation (LGHSR)  \cite{li2017large},  Discrete Spectral Hashing (DSH) \cite{hu2018discrete}, and Efficient Spectral Hashing (ESH) \cite{hemati2020non}. Other recent works on unsupervised hashing include,  Optimal Projection Hashing (OPH) \cite{chu2019optimized}
$k$-Nearest Neighbors Hashing (KNNH) \cite{He_2019_CVPR}, and Simultaneous Compression and Quantization (SCQ) \cite{HOANG2020102852}. Although deep learning has been mainly applied to the supervised hashing problem, some unsupervised deep hashing algorithms have recently been proposed. Some recent deep unsupervised hashing algorithms include deep hashing (DH) \cite{erin2015deep}, Unsupervised Hashing with Binary Deep Neural Network (UHBDNN) \cite{do2016learning}, Deepbit \cite{lin2016learning}, and Similarity-Adaptive Deep Hashing (SADH) \cite{shen2018unsupervised}.

Employing a neighbourhood preserving transformation used in many hashing algorithms. For example in the ITQ, IsoHash, AQ, and OPH a rotation matrix is employed to refine projections. Besides, neighbour preserving transformation has also been used in LGHSR and SCQ to improve the quality of SH relaxed solution and simultaneous projection and quantization of data respectively.

The main issue with binarization is destroying neighbourhood structure of data. Having this in mind,
rotation matrices seem to be a good choice for reducing quantization loss as they preserve neighbourhood. Although they are effective, we argue that a single rotation is not powerful enough to minimize the quantization error. Instead, we propose to employ transformations beyond rotation that even corrupt neighbourhood structure of data in favour pushing for quantization. As we will see, such transformation coupled with a rotation leads to high quality binary codes that outperforms state-of-art linear unsupervised hashing methods. Although the proposed idea can be applied to many hashing algorithms that employ a single rotation for reducing quantization error, here we choose to develop this idea on top of ITQ which is very fast and still among competitive hashing methods. In ITQ, the data is projected to lower dimensionality and then the optimal rotation is found such that it minimizes the quantization error in two separate steps.

In our method, we employ a transformation beyond rotation and also resolve the discontinuity in projecting data to lower dimensionality and reduce quantization error. We formulate and optimize our objective function such that in our algorithm:
\begin{enumerate}
  \item The data is projected and quantized simultaneously. This means unlike many conventional linear unsupervised hashing methods, the projection matrix is contributed towards reducing quantization error.
  \item We relax the orthogonality on projection matrix in favor of reducing quantization error even if it corrupts the original neighbourhood. We experiment different matrix norms for achieving this non-rigid transformation.
  \item An efficient sequential update scheme is  proposed for learning the projection matrix.
\end{enumerate}

The left plot in Fig. \ref{fig:1} shows the histogram of transformed data for a neighbourhood preserving transformation (ITQ) while the right shows the same data transformed by a non-neighbourhood preserving operations using our proposed method Sequential Non-rigid Quantization (SNRQ). Apparently,  SNRQ pushes data points more distinctively toward +1 and -1 modes compared with ITQ. 

Using extensive quantitative experiments on five public detest and also qualitative results, we show that although our transformation corrupts the neighbourhood of data, the final binary codes obtained using our method, preserve more neighbourhood compared with many other linear hashing methods.
\section{Proposed Method}
\textbf{Formulation -- } In this section, we formulate an optimization problem that follows two objectives. Firstly, it is desirable to jointly learn a projection matrix and minimize the quantization loss of projected data. Secondly, the orthogonality constraint on the projection matrix is relaxed such that we can employ a non-rigid transformation for reducing quantization error. These two objectives together enable non-orthogonal projection matrix to contribute to minimizing quantization loss in a way different from the orthogonal rotation matrix. The first objective is achieved by joining the two well known ITQ steps, namely applying PCA to the data and minimizing quantization loss. Let $\mathbf{X} \in \R ^{n\times D}$ represents zero-centered data in which $n$ is the number of training data points and $D$ is the data dimensionality. It is well understood that the projection matrix $\mathbf{W} \in \R ^{D\times K}$ or $K$ principal components of data can be obtained by maximizing the objective function
\begin{equation}
\underset{\mathbf{W}}{\text{arg max}} \quad \textrm{\emph{Tr}}\{\mathbf{W}^{T}\mathbf{X}^{T}\mathbf{X}\mathbf{W}\}\quad
\textrm{s.t.} \quad  \mathbf{W}^T \mathbf{W}=\mathbf{I}_K,
\label{eq:1}
\end{equation}
where $\mathbf{I}_K$ is the $K\times K$ identity matrix and \emph{Tr\{\}} is the trace of a matrix. Here we define $\mathbf{V}$ as $\mathbf{V}=\mathbf{X}\mathbf{W}$ with $\mathbf{V} \in \R ^{n\times K}.$ We would like to minimize the quantization loss of transformed data.
To do this, we find an orthogonality relaxed matrix $\mathbf{W}$ (the relaxed orthogonality constraint on $\mathbf{W}$ will be formulated later in Eq. \ref{eq:3}) and an $K\times K$ orthogonal rotation  matrix $\mathbf{R}$ such that the quantization loss of thresholding the transformed data  $\mathbf{V}\mathbf{R}=\mathbf{X}\mathbf{W}\mathbf{R}$ at zero is minimized. This can be formulated as minimizing the following:
\begin{equation}
Q(\mathbf{B},\mathbf{R},\mathbf{W})= \| \mathbf{X}\mathbf{W}\mathbf{R}-\mathbf{B} \| ^2_F \quad
\textrm{s.t.} \quad \mathbf{R}^T \mathbf{R}=\mathbf{I}_K, 
\label{eq:2}
\end{equation}
where $\mathbf{B} \in \{-1,1\} ^{n\times k}$ is the corresponding binary representation of $\mathbf{X}$. In order to jointly learn the projection matrix $\mathbf{W}$ and minimize the quantization loss using both non-rigid $\mathbf{W}$ and rigid $\mathbf{R}$, we relax the orthogonality constrain on the projection in PCA formulation and also regularize this by a quantization term which leads to \textbf{our proposed objective function}:
\begin{equation}
\label{eq:3}
\begin{split}
\underset{\mathbf{W}, \mathbf{R}, \mathbf{B}}{\text{arg max}}\quad \emph{J}(\mathbf{W}, \mathbf{R}, \mathbf{B})\!=\!
 \textrm{\emph{Tr}}\{\mathbf{W}^T\!\mathbf{C_x}\!\mathbf{W}\} - \\
 \alpha \| \mathbf{X}\mathbf{W}\mathbf{R}\!-\!\mathbf{B}\! \| ^2_F  - \beta \|\mathbf{W}^T \mathbf{W}\!-\!\mathbf{I}_K\|^2_F, 
\end{split}
\end{equation}
subject to $\mathbf{R}^T \mathbf{R}=\mathbf{I}_K$ and  $\mathbf{B} \in \{-1,1\} ^{n\times k}$ where $\alpha$ and $\beta$ are quantization and rigidness regularization parameters, and finally $\mathbf{C_x}=\mathbf{X}^{T}\mathbf{X}$. Note that in Eq. \ref{eq:3} both $\mathbf{W}$ and $\mathbf{R}$ are contributing to minimizing quantization loss, $\| \mathbf{X}\mathbf{W}\mathbf{R}-\!\mathbf{B}\! \| ^2_F$,  but in different ways as orthogonality constraint on $\mathbf{W}$ is smoothed whereas $\mathbf{R}$ is a pure orthogonal rotation matrix. This is one of the main differences between the proposed method and other existing methods e.g., in contrast to ITQ, IsoHash, AQ, and SCQ.

\textbf{Optimization -- } In order to jointly minimize the quantization loss and learn the projection matrix, we need to find $\mathbf{W}$, $\mathbf{R}$, and $\mathbf{B}$ in a way to maximize the objective function in Eq. \ref{eq:3}. Due to binary constraints on $\mathbf{B}$, the optimization is intractable. Hence, inspired by technique used in ITQ paper, we use a coordinate descent approach to optimize the objective function over $\mathbf{W}$, $\mathbf{R}$, and $\mathbf{B}$. To this end, in each step we consider one of the  $\mathbf{W}$, $\mathbf{R}$, and $\mathbf{B}$ variable and the two other matrices are assumed to be constant. 

\vspace{0.1in}
\noindent \textbf{Fix} $\mathbf{R}$ \textbf{and} $\mathbf{B}$\textbf{, and update} $\mathbf{W}$\textbf{:}
In order to optimize the objective in Eq. \ref{eq:3} with respect to $\mathbf{W}$, we easily calculate the derivative with respect to $\mathbf{W}$ as follow: 

\begin{equation}
\begin{split}
\frac{\partial \emph{J}(\mathbf{W}, \mathbf{R}, \mathbf{B}) }{\partial \mathbf{W}}=2\mathbf{\mathbf{C_x}}\mathbf{W}-\alpha \left(2 \mathbf{X}^{T}(\mathbf{X}\mathbf{W}\mathbf{R}-\mathbf{B}) \mathbf{R}^T \right) \\
-\beta \left(4\mathbf{W}(\mathbf{W}^T\mathbf{W}-\mathbf{I}_K) \right).
\end{split}
\label{eq:4}
\end{equation}

Having the gradient w.r.t $\mathbf{W}$, we found that the L-BFGS-B optimizer \cite{zhu1997algorithm} obtain a good solution for $\mathbf{W}$. However, as each variable should be updated multiple times (while other are fixed) we observed that optimizing for $\mathbf{W}$ directly is a time-consuming procedure. To mitigate this challenge, in the following we propose a sequential approach for updating $\mathbf{W}$. As we will show in experiments, sequential approach achieves comparable performance while reducing training time significantly. In sequential scheme, each time one column of $\mathbf{W}$ is updated while  the rest of columns are considered fixed. First, let's expand the second term (quantization loss) in Eq. \ref{eq:3}:
\begin{equation}
\begin{aligned}
&\| \mathbf{X}\mathbf{W}\mathbf{R}-\mathbf{B} \| ^2_F=\|\mathbf{X}\mathbf{W}\mathbf{R}\|^2_F-2\textrm{\emph{Tr}}\{\mathbf{X}\mathbf{W}\mathbf{R}\mathbf{B}^T\}+\|\mathbf{B}\|^2_F \\
=&\textrm{\emph{Tr}}\{\mathbf{W}^T\mathbf{C_x}\mathbf{W}\}-2\textrm{\emph{Tr}}\{\mathbf{W}(\mathbf{R}\mathbf{B}^T\mathbf{X})\}+const.
\end{aligned}
\label{eq:5}
\end{equation}

Now let's expand the relaxed orthogonality constraint (the third term in Eq. \ref{eq:3}):
\begin{equation}
\label{eq:6}
\begin{aligned}
&\|\mathbf{W}^T \mathbf{W}- \mathbf{I}_K\|^2_F=\|\mathbf{W}^T \mathbf{W}\|^2_F-2 \textrm{\emph{Tr}}\{\mathbf{W}^T \mathbf{W}\mathbf{I}_K\}+const\\
=&\textrm{\emph{Tr}}\{(\mathbf{W} \mathbf{W}^T)(\mathbf{W} \mathbf{W}^T)\}-2 \textrm{\emph{Tr}}\{\mathbf{W} \mathbf{W}^T\}+const.
\end{aligned}
\end{equation}

Using Eqs. \ref{eq:5} and \ref{eq:6} for the second and third terms in Eq. \ref{eq:3}, respectively, and defining  $\mathbf{C_y}=\mathbf{R}\mathbf{B}^T\mathbf{X}$ provides us with the objective function for the case that $\mathbf{W}$ is variable:

\begin{equation}
\begin{aligned}
& \underset{\mathbf{W}}{\text{arg max}}\quad (1-\alpha)\textrm{\emph{Tr}}\{\mathbf{W}^{T}\mathbf{C_x}\mathbf{W}\}+2\alpha \textrm{\emph{Tr}}\{\mathbf{W}\mathbf{C_y}\} \\
& -\beta \textrm{\emph{Tr}}\{(\mathbf{W} \mathbf{W}^T)(\mathbf{W} \mathbf{W}^T)\}+2\beta \textrm{\emph{Tr}}\{\mathbf{W}\mathbf{W}^T\}.
\label{eq:7}
\end{aligned}
\end{equation}
Let's start formulating the problem such that one column of $\mathbf{W}$ is considered variable while the remaining columns are constant. To this end, the $k$-th column of $\mathbf{W}$ that is considered variable is denoted by $\mathbf{z}_k$ and all other columns are shown by $\mathbf{W'}$. In this case, for the first term we receive
\begin{equation}
\begin{aligned}
&\textrm{\emph{Tr}}\{\mathbf{W}^{T}\mathbf{C_x}\mathbf{W}\}=\textrm{\emph{Tr}}\{\mathbf{C_x}\mathbf{W}\mathbf{W}^{T}\}=\\
& \textrm{\emph{Tr}}\{\mathbf{C_x}(\mathbf{W'} \mathbf{W'}^T+ \mathbf{z}_k  \mathbf{z}_k^T)\}=const+\mathbf{z}_k^T\mathbf{C_x} \mathbf{z}_k.
\end{aligned}
\label{eq:8}
\end{equation}
Note that here first we used the fact that $\textrm{\emph{Tr}}\{\mathbf{C_x} \mathbf{z}_k \mathbf{z}_k^T\}= \textrm{\emph{Tr}}\{ \mathbf{z}_k^T \mathbf{C_x} \mathbf{z}_k\}$ and then removed $\textrm{\emph{Tr}}\{\cdot\}$ as the $\mathbf{z}_k^T \mathbf{C_x} \mathbf{z}_k$ is scalar.
Similarly, if we define the $k$-th row of $\mathbf{C_y}$ as $\mathbf{u}_k^T$ and the rest of the rows as $\mathbf{C'_y}$, then the second term can be written as
\begin{equation}
\begin{aligned}
\textrm{\emph{Tr}}\{\mathbf{W}\mathbf{C_y}\}\!=\!\textrm{\emph{Tr}}\{\mathbf{W'} \mathbf{C'_y}+\mathbf{z}_k\mathbf{u}_k^T\}\!=\\
const+\!\textrm{\emph{Tr}}\{\mathbf{u}_k^T \mathbf{z}_k\}\!=const+\!\mathbf{u}_k^T\mathbf{z}_k.
\end{aligned}
\label{eq:9}
\end{equation}

For the third term in Eq. \ref{eq:7} we write

\begin{equation}
\begin{aligned}
&\textrm{\emph{Tr}}\{(\mathbf{W} \mathbf{W}^T)(\mathbf{W} \mathbf{W}^T)\}=\\
&\textrm{\emph{Tr}}\{(\mathbf{W'} \mathbf{W'}^T+ \mathbf{z}_k  \mathbf{z}_k^T)(\mathbf{W'} \mathbf{W'}^T+ \mathbf{z}_k  \mathbf{z}_k^T)\}=\\
& const + 2 \mathbf{z}_k^T \mathbf{W'} \mathbf{W'}^T \mathbf{z}_k+ \mathbf{z}_k^T \mathbf{z}_k\mathbf{z}_k^T\mathbf{z}_k.
\end{aligned}
\label{eq:10}
\end{equation} 
As $\mathbf{z}_k^T \mathbf{W'} \mathbf{W'}^T \mathbf{z}_k$ and $\mathbf{z}_k^T \mathbf{z}_k\mathbf{z}_k^T\mathbf{z}_k$ are scalars, $\textrm{\emph{Tr}}\{\cdot\}$ can be removed.
Similarly, for the last term we obtain
\begin{equation}
\begin{aligned}
\textrm{\emph{Tr}}\{\mathbf{W}\mathbf{W}^T\}\!=\!\textrm{\emph{Tr}}\{\mathbf{W'} \mathbf{W'}^T +\mathbf{z}_k \mathbf{z}_k^T\}\!=\\
const + \!\textrm{\emph{Tr}}\{\mathbf{z}_k^T\mathbf{z}_k\} \!=\! const + \mathbf{z}_k^T\mathbf{z}_k.
\end{aligned}
\label{eq:11}
\end{equation}

Incorporating Eqs. \ref{eq:8}, \ref{eq:9}, \ref{eq:10}, and \ref{eq:11} into Eq. \ref{eq:7}, if we set
\begin{equation}
\mathbf{Q}=(1-\alpha)\mathbf{C_x}-2\beta \mathbf{W'} \mathbf{W'}^T+2\beta \mathbf{I}_D,
\label{eq:12}
\end{equation} then by maximizing the following objective function one can obtain the $\mathbf{z}_k$, $k$-th column of $\mathbf{W}$:
\begin{equation}
\underset{\mathbf{z}_k}{\text{arg max}}\quad \emph{J}(\mathbf{z}_k)\!=\! \mathbf{z}_k^T \mathbf{Q}\mathbf{z}_k\!+\!2 \alpha \mathbf{u}_k^T\mathbf{z}_k\!-\!\beta \mathbf{z}_k^T \mathbf{z}_k\mathbf{z}_k^T\mathbf{z}_k.
\label{eq:13}
\end{equation}

To update each column of $\mathbf{W}$, i.e, $\mathbf{z}_k$, where $k=1 \dots K,$ we have to find $\mathbf{z}_k$ that maximizes Eq. \ref{eq:13}. The gradient of objective function (Eq. \ref{eq:13}) can be calculated as
\begin{equation}
\frac{\partial \emph{J}(\mathbf{z}_k) }{\partial \mathbf{z}_k}=
2 \mathbf{Q}\mathbf{z}_k+2\alpha \mathbf{u}_k-4\beta \mathbf{z}_k (\mathbf{z}_k\mathbf{z}_k^T).
\label{eq:14}
\end{equation}
Now, we use the L-BFGS-B optimizer to obtain solutions. As we will show in experiments, this sequential learning scheme significantly reduces the training time of the proposed algorithm without sacrificing the quality of binary codes. This efficiency is in part because of simplifying the objective function and also reducing the search space by dividing the problem to $K$ sub-problems.

\vspace{0.15in}
\noindent \textbf{Fix} $\mathbf{R}$ \textbf{and} $\mathbf{W}$\textbf{, and update} $\mathbf{B}$\textbf{:} In this case, the optimization in Eq. \ref{eq:3} takes the following form:

\begin{equation}
\begin{aligned}
& \underset{\mathbf{B}}{\text{arg min}} &&   \| \mathbf{X}\mathbf{W}\mathbf{R}-\mathbf{B} \| ^2_F  \quad 
\textrm{s.t.} \quad  \mathbf{B} \in \{-1,1\} ^{n\times k}.
\label{eq:15}
\end{aligned}
\end{equation}

Having in mind that $\mathbf{R}$ and $\mathbf{W}$ are fixed, 
clearly, minimization of Eq. \ref{eq:15} with respect to $\mathbf{B}$ is equivalent the maximization of $\textrm{\emph{Tr}}\{(\mathbf{X}\mathbf{W}\mathbf{R})^T\mathbf{B}\}$ where elements of $\mathbf{B}$ can be either 1 or -1. As we know from ITQ  \cite{gong2012iterative}, the optimal $\mathbf{B}$ can be calculated as
\begin{equation}
\mathbf{B}=sgn(\mathbf{V}\mathbf{R})=sgn(\mathbf{X}\mathbf{W}\mathbf{R}).
\label{eq:16}
\end{equation}

\vspace{0.15in}
\noindent \textbf{Fix} $\mathbf{W}$ \textbf{and} $\mathbf{B}$\textbf{, and update} $\mathbf{R}$\textbf{:} For the case that $\mathbf{R}$ is variable, maximizing Eq. \ref{eq:3} is equivalent to

\begin{equation}
\begin{aligned}
& \underset{\mathbf{R}}{\text{arg min}} &&  
\|\mathbf{V}\mathbf{R}-\mathbf{B} \| ^2_F \quad
\textrm{s.t.} \quad \mathbf{R}^T \mathbf{R}=\mathbf{I}_K.
\label{eq:17}
\end{aligned}
\end{equation}

This is the Orthogonal Procrustes problem where a rotation matrix is found such that two point sets are aligned with each other. Here, these two point sets are the target binary code matrix $\mathbf{B}$ and projected data $\mathbf{V}$. This problem has closed form solution when $\mathbf{R}$ is a square orthogonal (rotation) matrix \cite{schonemann1966generalized} which is

\begin{equation}
\mathbf{R}=\hat{{\mathbf{S}}}{\mathbf{S}}^T.
\label{eq:18}
\end{equation}
where ${\mathbf{S}} \mathbf{\Lambda}  \hat{{\mathbf{S}}}^T$ is the singular value decomposition (SVD) of the $K$ by $K$ matrix $\mathbf{B}^T\mathbf{V}$.

In summary, in each iteration, Eq. \ref{eq:16} and Eq. \ref{eq:18} are used to update $\mathbf{B}$ and $\mathbf{R}$, respectively.To update $\mathbf{W}$, in each iteration, $K$ optimization problems for $K$ columns of $\mathbf{W}$ are solved. After obtaining $\mathbf{R}$, and $\mathbf{W}$, the binary representation of the data $\mathbf{X}$ can be obtained $\mathbf{B}=sgn(\mathbf{V}\mathbf{R})=sgn(\mathbf{X}\mathbf{W}\mathbf{R})$.  When the direct optimization is used for updating $\mathbf{W}$ we denote our method Non-rigid Quantization (NRQ) and for the sequential case this denoted by SNRQ.
Algorithm \ref{alg1} summarizes the proposed scheme to update all three matrices method.

\textbf{Implementation Note -- }
We use $K$ eigenvectors of $\mathbf{C}_x$ corresponding to $K$ largest eigenvalues as initialization of $\mathbf{W}$. For $\mathbf{R}$, we use the rotation matrix obtained by ITQ as initialization of $\mathbf{R}$. Although we will show the algorithm is robust to different values for $\alpha$ and $\beta$, to avoid tuning regularization parameters, we set $\alpha=3$ and $\beta=0.01$ for all datasets and experiments. 
In order to choose $\alpha$, the only consideration is to keep it larger than 1. This can be understood based on Eq. \ref{eq:12} which shows there is a trade-off between reducing quantization error and maximizing variance across projections. This is clear that the quantization error is our priority. The algorithm is fairly robust to $\beta$, and based on our experiments, any number between $0.01$ to $50$ easily does the job. For the number of iterations, experiments on a variety of detests show that 70 iterations are generally sufficient, and after that, the loss function does not change significantly. As a results, We set the number of iterations $N$ to 70. An implementation of the proposed method is provided in supplementary material.

\begin{algorithm}[t]
	\caption{The Proposed SNRQ Algorithm} 
	
	\hspace*{\algorithmicindent} \textbf{Input:} Data matrix \textbf{X}, number of iterations $N$\\
	\hspace*{\algorithmicindent} regularization parameters $\alpha$ and $\beta$ \\
    \hspace*{\algorithmicindent} \textbf{Output:} Projection matrix $\mathbf{W}$, rotation matrix $\mathbf{R}$ \\ 
    \hspace*{\algorithmicindent} \textbf{Initialization:} Initialize $\mathbf{W}$ and $\mathbf{R}$
	\begin{algorithmic}[1]
		\For {iteration$=1,2,\ldots,N$}
		\State $\mathbf{V} \leftarrow \mathbf{X}\mathbf{W}$
		\State $\mathbf{B}\leftarrow sgn(\mathbf{V}\mathbf{R})$
	    \State $\textbf{S} \Lambda \hat{{\textbf{S}}}^T \leftarrow    \textrm{SVD}(\textbf{B}^T\mathbf{V})$
	    \State $\textbf{R} \leftarrow \hat{{\textbf{S}}}{\textbf{S}}^T$ 
		\State $\mathbf{C_y}\leftarrow \mathbf{R}\mathbf{B}^T\mathbf{X}$
		
			\For {$k=1,2,\ldots,K$}
			    \State $\mathbf{u}_k^T \leftarrow \mathbf{C_y}[k,:]$
			      \State $\mathbf{Q}\leftarrow(1-\alpha)\mathbf{C_x}-2\beta \mathbf{W'} \mathbf{W'}^T+2\beta  \mathbf{I}_K$
				\State Solve Eq. $\ref{eq:13}$ for each  column of $\mathbf{W}$, i.e., $\mathbf{z}_k$ 
				\State $\emph{J}(\mathbf{z}_k)= \mathbf{z}_k^T \mathbf{Q}\mathbf{z}_k+2 \alpha \mathbf{u}_k^T\mathbf{z}_k-\beta \mathbf{z}_k^T \mathbf{z}_k\mathbf{z}_k^T\mathbf{z}_k$ 
				\State $\frac{\partial \emph{J}(\mathbf{z}_k) }{\partial \mathbf{z}_k}=2 \mathbf{Q}\mathbf{z}_k+2\alpha \mathbf{u}_k-4\beta \mathbf{z}_k (\mathbf{z}_k\mathbf{z}_k^T)$
				\State Update the $k$-th column of $\mathbf{W}$, i.e., $\mathbf{z}_k$ 
				\State $\mathbf{W}[:,k] \leftarrow \mathbf{z}_k$
			\EndFor
		\EndFor
	\end{algorithmic} 
\label{alg1}
\end{algorithm}

\section{Experiments and Results}
\textbf{Datasets and Evaluation protocol}
The performance of the proposed  SNRQ algorithm is evaluated on five standard benchmark datasets,  MNIST \cite{lecun1998gradient}, CIFAR-10 \cite{krizhevsky2009learning},  an unbalanced dataset LabelMe-12-50K \cite{uetz2009large}, a medical image dataset NCT-CRC-HE-100K \cite{macenko2009method}, and a multi-label NUS-WIDE \cite{10.1145/1646396.1646452}. These datasets provide a total of \textbf{489,000 images} for learning and testing.  

\vspace{0.1in}
The \textbf{\underline{MNIST dataset}} contains 70,000 gray-scale images all of size $28\times 28$ pixels. There are 10 classes in this dataset for handwritten digits. 

The \textbf{\underline{CIFAR-10 dataset}} is a 10-class dataset consisting of 60,000 color images of size $32 \times 32$ pixels.  

The \textbf{\underline{LabelMe-12-50K dataset}} is a 12-class dataset containing 50,000 images of size $256\times 256$ pixels. This dataset is highly imbalanced such that five classes constitute 91\% of all images while there is one class that only contains 0.6\% of the samples. The images of this dataset have multiple label values  between zero and one. In our experiments, same as previous works that employed this dataset  \cite{He_2019_CVPR} for evaluating hashing algorithms, we choose the class of the largest label value as the image label.

The \textbf{\underline{NCT-CRC-HE-100K dataset}} is 9-class histopathology dataset containing 100,000 non-overlapping image patches from hematoxylin \& eosin stained (H\&E)  images of human colorectal cancer and normal tissue. All images are $224\times 224$ pixels and color-normalized. 

The \textbf{\underline{NUS-WIDE dataset}} is a multilabel dataset that contains 269,000 images collected from Flickr. This database contains 81 ground-truth concepts. 

\textbf{Hash Code Evaluation}
To evaluate the performance of the SNRQ, we use standard measures for image retrieval quality assessment. These measures include mean Average Precision (mAP), and precision at $M$ samples (e.g., precision@1000). Briefly, mAP measures the overall performance of the retrieval over all classes, whereas precision@$M$ calculates the proportion of true positive over top $M$ retrieved samples. 
\begin{table}[!htp]
\centering
\footnotesize
\caption{Comparison of retrieval performance based on mAP  and precision@1000 on MNIST dataset represented by 512-D GIST descriptor. The best performance is highlighted in boldface. KMH: K-means Hashing \cite{he2013k},   SpH: spherical hashing \cite{heo2012spherical}.}
\begin{tabular}{|l|ccc|ccc|}
\hline
                  & \multicolumn{3}{c|}{mAP \%}    & \multicolumn{3}{c|}{precision \% @1000}                                                                                                                                           \\ \hline
Method                    & \multicolumn{1}{c|}{16 bits}       & \multicolumn{1}{c|}{32 bits}       & \multicolumn{1}{c|}{64 bits}      & \multicolumn{1}{c|}{16 bits}  & \multicolumn{1}{c|}{32 bits}       & \multicolumn{1}{c|}{64 bits}               \\ \hline\hline  
 SH                       & 32.59                                   &33.23                                &30.65                                   & -    & -    & -                                                      \\ \hline
 SpH                        & 31.27                                   & 36.80                                    & 41.40                                   & -    & -   & -                                                     \\ \hline
KMH                           & 31.96                                    &37.39                                    &41.11                                   & -    
& -   & -                                                 \\ \hline
BA                         & 48.48                                  &51.72                                & 52.73                                  & -      & -    & -    -                                                    \\ \hline

ITQ             & 46.37                             & 50.59                             & 53.69                              & 69.67       & 75.13       & 80.45                                          \\ \hline
KNNH           & 53.07                              & 61.11                             & 65.55                              &73.99          &83.32  &87.05                                       \\ \hline                                                       

SCQ         & 62.39                      &74.49                   &72.23          & 79.26                        &88.46      & \textbf{88.90}      
\\ \hline   

NRQ (Ours)  &\textbf{71.46}     & 69.44  &\textbf{72.79} & \textbf{84.59} & 85.49 &87.30 

\\ \hline

SNRQ (Ours)            &64.70                            & \textbf{76.98}                              & \textbf{73.48}                           & 80.00         & \textbf{88.53}       &88.27                  \\ \hline
\end{tabular}
\label{table:table1}
\end{table}

\begin{table}[!htb]
\centering
\footnotesize
\caption{Comparison of retrieval performance, based on mAP, for 16, 32 and 64 bits  for CIFAR-10 and macro mAP (average over classes) for LabelMe-12-50k datasets both represented by 4096-D VGG-FC7 descriptors.}
\begin{tabular}{|l|ccc|ccc|}
\hline
                  & \multicolumn{3}{c|}{CIFAR-10}                     & \multicolumn{3}{c|}{LabelMe-12-50k}                                                                      \\ \hline                                                                                   
Method                    & \multicolumn{1}{c|}{16 bits}       & \multicolumn{1}{c|}{32 bits}       & \multicolumn{1}{c|}{64 bits}     & \multicolumn{1}{c|}{16 bits}       & \multicolumn{1}{c|}{32 bits}       & \multicolumn{1}{c|}{64 bits}          \\ \hline\hline                             
 SH                       & 18.31                                  &16.54                             &15.78            & 12.60                & 12.59           & 12.24                                                           \\ \hline
 
 SpH                      & 18.82                            &20.93                                   & 23.40    & 13.59                  &15.10         &17.03                                                                                                       \\ \hline
KMH                         & 18.68                                   &20.82                       &22.87                    &13.36            &15.47                 &16.58                                                                             \\ \hline

BA                        & 25.38                         &26.16                             & 27.99             & 16.96         &18.42       &20.80                                        \\ \hline

ITQ            &26.82                            & 27.38                              & 28.73         & 18.06              & 19.40                &20.73                                                         \\ \hline

DGH     & 27.73         & 27.44                            &28.01    &21.45    	&22.74	   &25.41   	\\ \hline  

LGHSR  &27.83  &25.87	&22.12 &21.10	&23.49	&23.98	
\\\hline 
DSH      &27.72	              & 25.36                             &22.12  &24.70	 &23.78	   &24.35	   
 \\ \hline

 SCQ           & 27.52                        &27.42                    &30.34          & 22.89                        &24.95      &26.50    \\ \hline 
KNNH           & 29.06                          &  30.82                               &32.60        &20.13           &23.79                      &26.22                                                               
                                                           \\ \hline

NRQ (Ours)        & \textbf{31.42}                            &30.87                              &33.05      &24.27                            &\textbf{27.40}                              &\textbf{28.65}   \\ \hline 

SNRQ (Ours)        &30.10                            & \textbf{31.82}                              & \textbf{33.10}      &\textbf{25.14}                            & 26.80                              & 28.50                                                                  \\ \hline

\end{tabular}
\label{table:table2}
\end{table}
\textbf{Results on MNIST dataset:}

Following the setting of \cite{He_2019_CVPR} for MNIST data, each image is presented by a GIST 512-D descriptor, 10\% of each class is considered for query set and the remaining data is used as training set. Table \ref{table:table1} shows the results for MNIST in terms of mAP,  and precision@1000. Clearly, the NRQ and SNRQ methods provide an improvement over other methods in all cases with only one exception for precision@1000 where the SCQ slightly outperforms SNRQ with less than 0.5\% in 64 bit.

\textbf{Results on CIFAR-10 dataset:} For this dataset, following the common setting \cite{He_2019_CVPR, zhang2015bit}, we used deep 4096-D features extracted from VGG network \cite{simonyan2014very} and sample 10\% of each class as query set and the remaining instances as training set. 
The left half of Table \ref{table:table2} represents the results for this experiment setting. As it can be seen, NRQ and SNRQ outperform state-of-art namely KNNH \cite{He_2019_CVPR} and SCQ \cite{HOANG2020102852}.

\textbf{Results on LabelMe-12-50k dataset:} As it was pointed out, this dataset is highly imbalanced. To ascertain a fair comparison, we calculate mAP values that are macro averages over all classes. Following the common setting \cite{He_2019_CVPR} we sample 10\% of each class to construct the query set, and the remaining data points as training set. Besides, VGG network has been used to extract feature vectors.  As Table \ref{table:table2} shows, NRQ and SNRQ are performing better and outperform state-of-art methods with a large gap.

\textbf{Results on NCT-CRC-HE-100K dataset:} For this dataset, we used EfficientNet \cite{tan2019efficientnet} pre-trained on ImageNet to extract 1280-D feature vectors from images. We randomly sampled 70,000 images (out of 100K) for training set and the rest of images (30,000) for test. For efficiency, we reduced the feature vector dimensionality to 512 values by the PCA.  Table \ref{table:table3} shows that NRQ and SNRQ outperform competitive methods based on mAP and precision@1000 with a significant gap. 

\begin{table}[]
\centering
\footnotesize
\caption{Comparison of retrieval performance based on mAP, and precision@1000 on NCT-CRC-HE-100K dataset represented by features extracted by EfficientNet.}
\begin{tabular}{|l|ccc|ccc|}
\hline
                  & \multicolumn{3}{c|}{mAP \%}    & \multicolumn{3}{c|}{precision \% @1000}                                                                                                                                                                     \\ \hline
Method                    & \multicolumn{1}{c|}{16 bits}       & \multicolumn{1}{c|}{32 bits}       & \multicolumn{1}{c|}{64 bits}      & \multicolumn{1}{c|}{16 bits}  & \multicolumn{1}{c|}{32 bits}       & \multicolumn{1}{c|}{64 bits}           \\ \hline\hline  

ITQ             & 55.8                             & 57.8                            & 59.5                              & 68.8      & 72.6      & 75.4                                       \\ \hline

DGH 
  &49.85	 &47.24	   &51.61	   	   &74.08	 &77.14	   &77.05	  

\\ \hline

LGHSR   &55.28	&47.74	&38.65		&75.66	&78.21	&77.13	
\\ \hline

DSH   &58.75	&45.63	&37.27	 &76.70	 &78.71	&77.73	 		
\\ \hline

KNNH             &58.02	  &61.47	  &64.28	&70.27	  &74.91	&78.16	 	                         \\ \hline                                        
SCQ             & 64.90                         & 67.25                             &   67.75                      & 76.03      &80.07      & 81.01 
\\ \hline

NRQ (Ours)             & \textbf{75.32}                            &73.08                              &77.25                          &\textbf{83.94}         &83.32       &86.91     \\ \hline

SNRQ  (Ours)            &68.5                           & \textbf{75.4}                              & \textbf{78.7}                           & 80.1        &\textbf{84.5}       &\textbf{87.0}             \\ \hline

\end{tabular}
\label{table:table3}
\end{table}

\textbf{Results on NUS-WIDE dataset:}
For NUS-WIDE experiments, we used the common setup in many hashing papers \cite{liu2011hashing, shen2018unsupervised}. In this setting, images are selected which their labels are among the 21 most frequent labels. This leads to 195,834 images. We randomly sample 2,100 images (100 images from each class) from 195,834 images for the test set and the rest of the images are used for training the hash function and populating the hash table. We used VGG-F network \cite{BMVC.28.6} to extract features from images. Results for this experiment are reported in Table \ref{table:table4}.  The results for other algorithms mainly are directly reported from literature. \cite{shen2018unsupervised}. 

\begin{table}[]
\centering
\small
\footnotesize
\caption{Comparison of retrieval performance on NUS-WIDE dataset represented by VGG-F deep features.}
\begin{tabular}{|l|ccc|ccc|}
\hline
                  & \multicolumn{3}{c|}{mAP \%}    & \multicolumn{3}{c|}{precision \% @5000}                                                                                                                                                                      \\ \hline
Method                    & \multicolumn{1}{c|}{16 bits}       & \multicolumn{1}{c|}{32 bits}       & \multicolumn{1}{c|}{64 bits}      & \multicolumn{1}{c|}{16 bits}  & \multicolumn{1}{c|}{32 bits}       & \multicolumn{1}{c|}{64 bits}             \\ \hline\hline  
                                                  
 SH                      &44.74                                    &42.60                                &42.36                                 & 59.15    & 54.98    & 54.18                                                    \\ \hline

AGH                        & 49.80        &47.14                             & 44.72                                  & 70.43    & 70.29   & 69.29  
                                              \\ \hline
DGH                     &54.03                                   &52.74                                & 49.64                                  & \textbf{71.15}      & 71.91    & 70.66                                                     \\ \hline

LGHSR           & 50.79                            & 47.72                           &  45.45                          &66.79           & 68.59           &67.60                                           \\ \hline
ITQ            &  53.39                           &   54.61                         & 55.75                            & 65.80          &   68.69         
& 70.59                                          \\ \hline

DSH        &49.87	&47.07	&45.67		&68.05	&68.17	&67.83		    
  \\ \hline                                                        
KNNH    &55.12	&57.53	&58.61		&66.77	&69.83	&71.07		
           \\ \hline                                                       
SCQ           &  58.34                           &  56.21                        &  56.10                          & 68.45          & 70.64         
& 70.81                                      \\ \hline

NRQ (Ours)             &60.83                            &62.08                          &\textbf{63.34}                          &69.91       &72.07          &\textbf{73.82}                \\ \hline

SNRQ (Ours)             &\textbf{61.78}                            & \textbf{62.74}                              &62.60                           & 70.83        &\textbf{72.33}          & 73.37                \\ \hline

\end{tabular}
\label{table:table4}
\end{table}

\begin{figure}[!htp]%
    \centering
\includegraphics[width=.75\columnwidth]{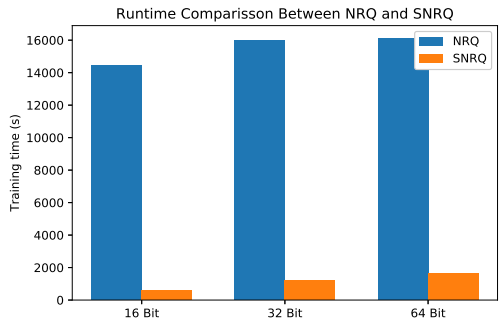}\\
\caption{Training time for NRQ and SNRQ applied to the CIFAR-10 dataset for different number of bits.}%
    \label{fig:2}%
\end{figure}

\begin{figure*}[!htp]
  \centering

  \medskip

  \begin{subfigure}[t]{.3\linewidth}
    \centering\includegraphics[width=4cm]{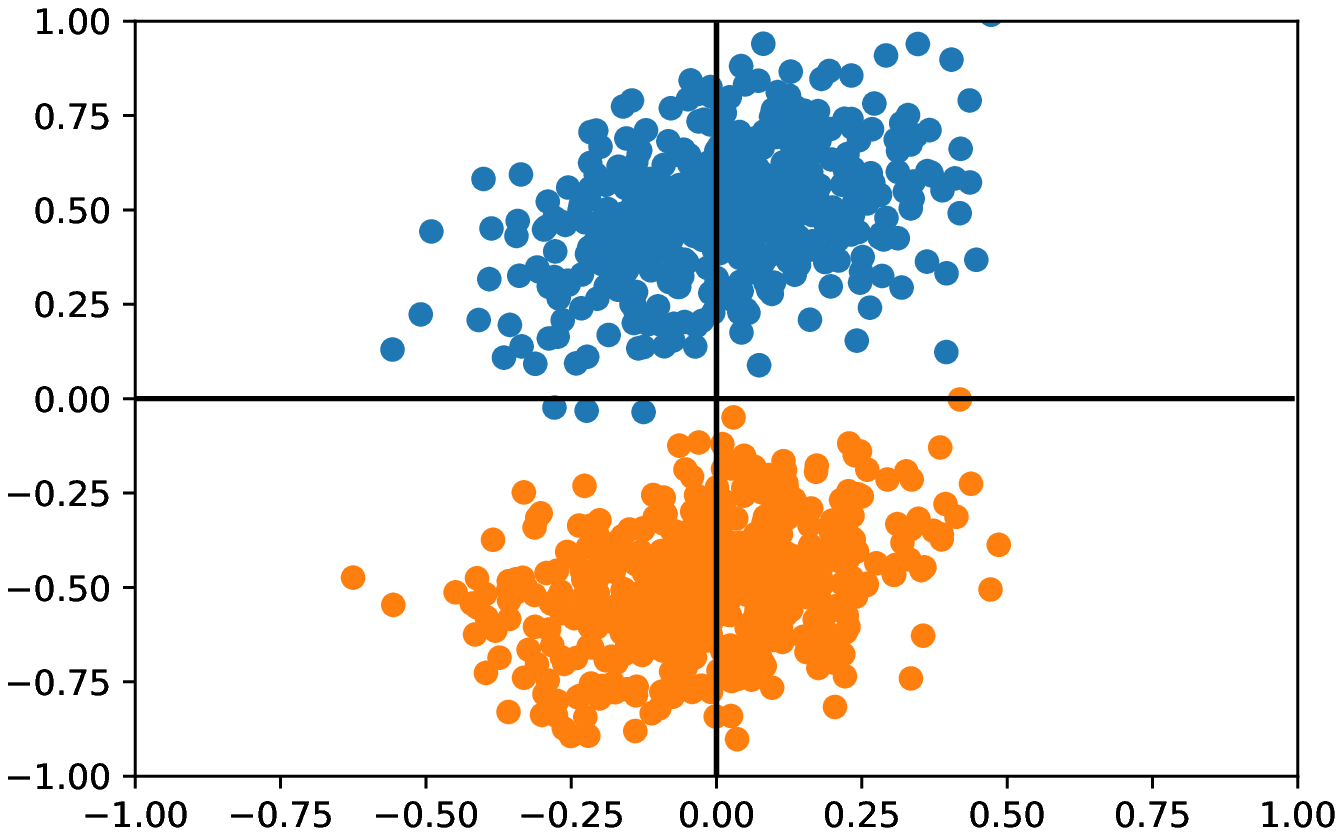}
    
    \caption{Synthetic toy data}
  \end{subfigure}
  \begin{subfigure}[t]{.3\linewidth}
    \centering\includegraphics[width=4cm]{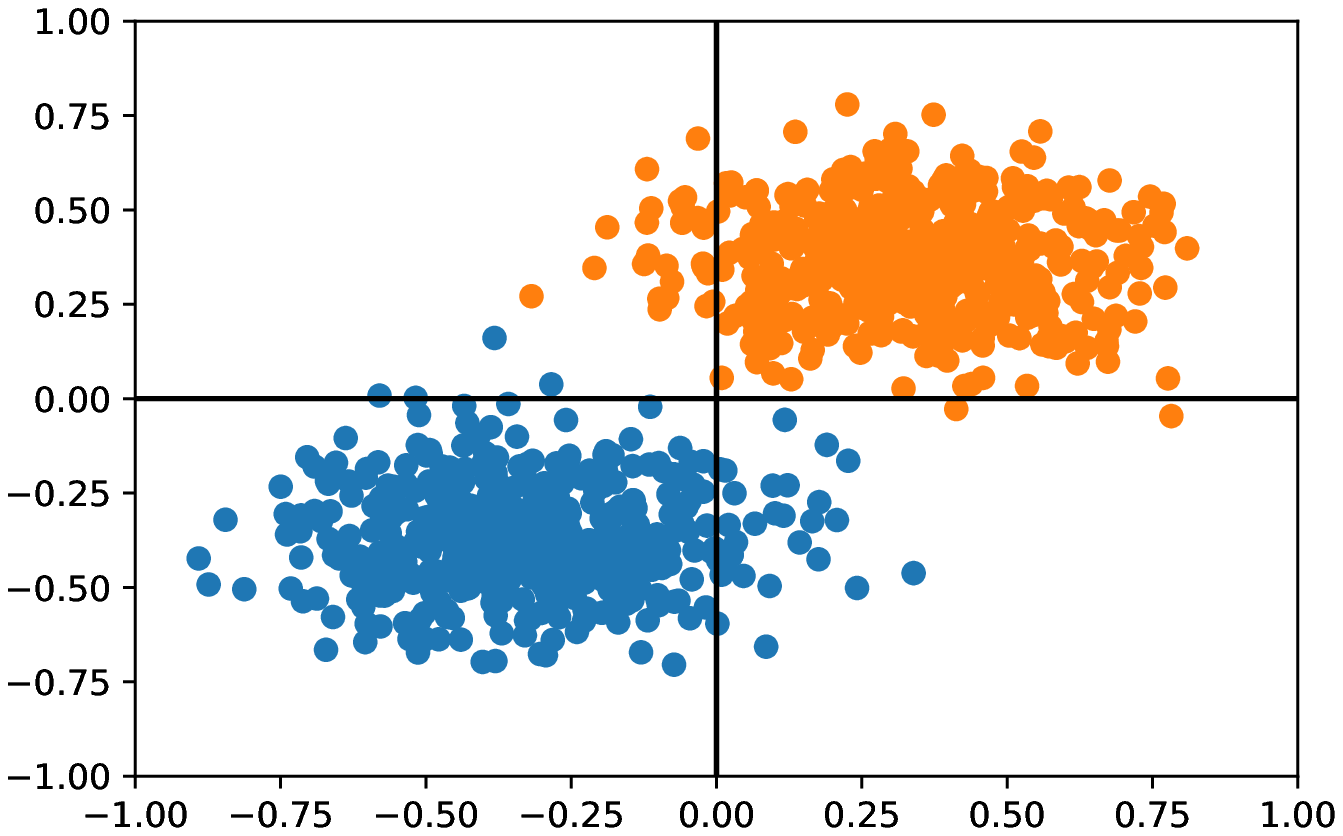}
    \caption{ITQ}
  \end{subfigure}
    \begin{subfigure}[t]{.3\linewidth}
    \centering\includegraphics[width=4cm]{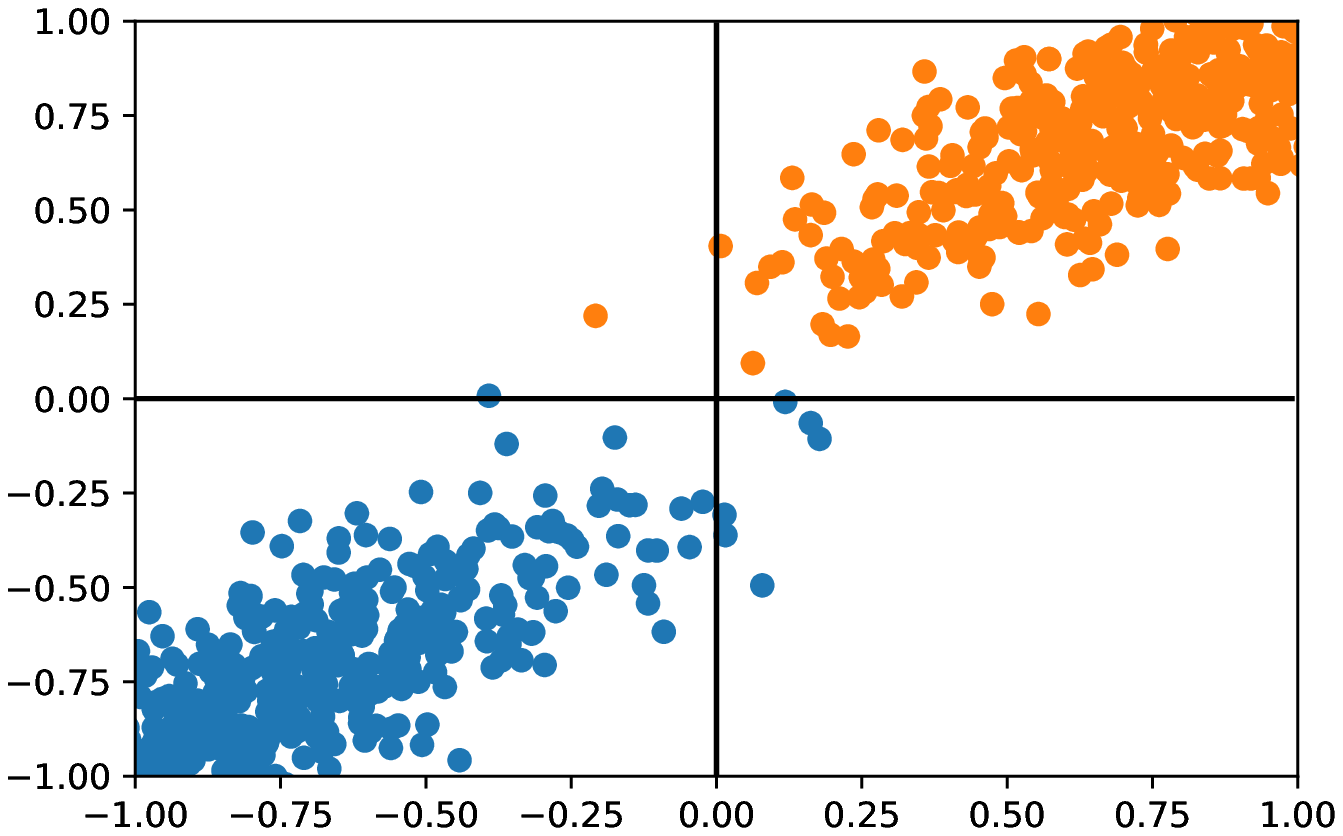}
    \caption{SNRQ ($\alpha\!=\!3$.\,  $\beta\!=\!50$)}
    \end{subfigure}
    \begin{subfigure}[t]{.3\linewidth}
    \centering\includegraphics[width=4cm]{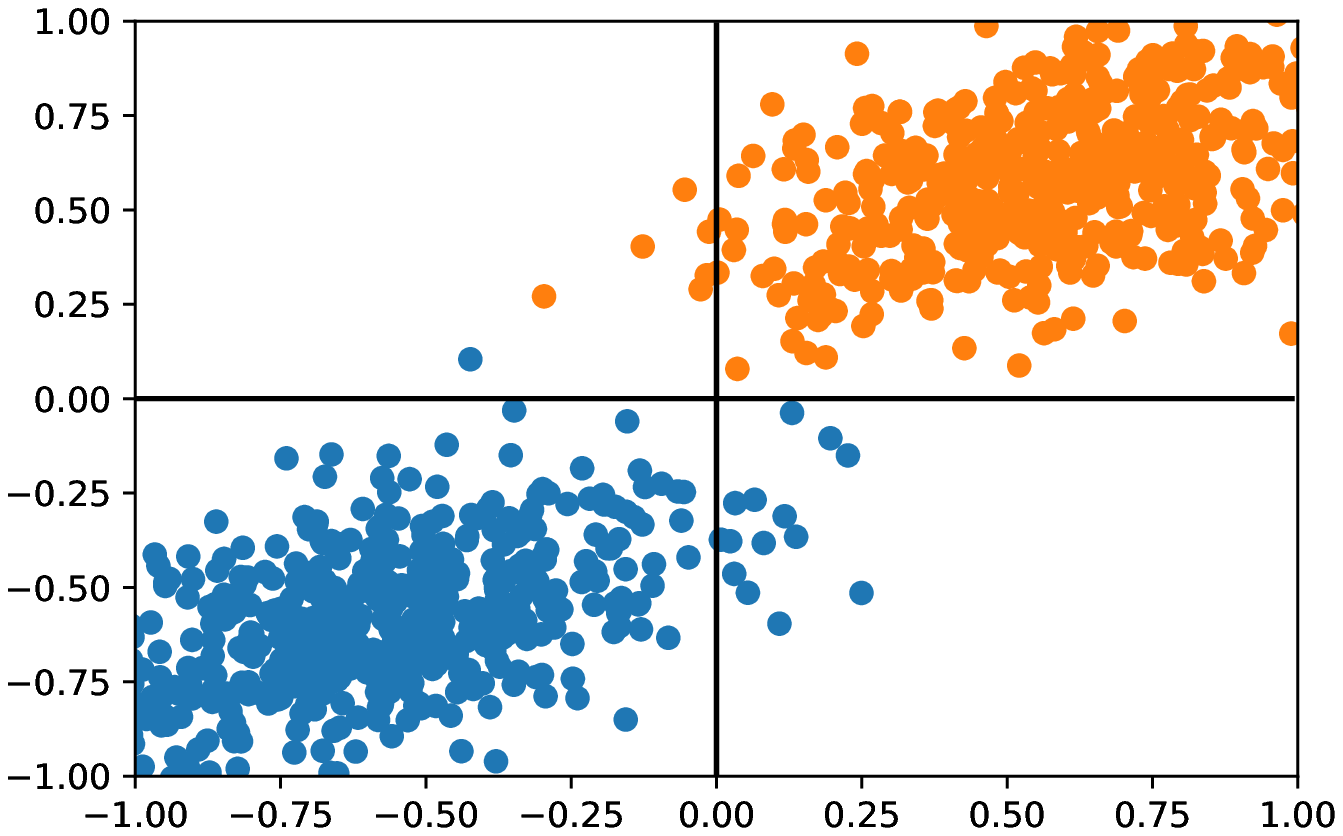}
    \caption{SNRQ ($\alpha\!=\!3$,  $\beta\!=\!200$)}
    \end{subfigure}
    \begin{subfigure}[t]{.3\linewidth}
    \centering\includegraphics[width=4cm]{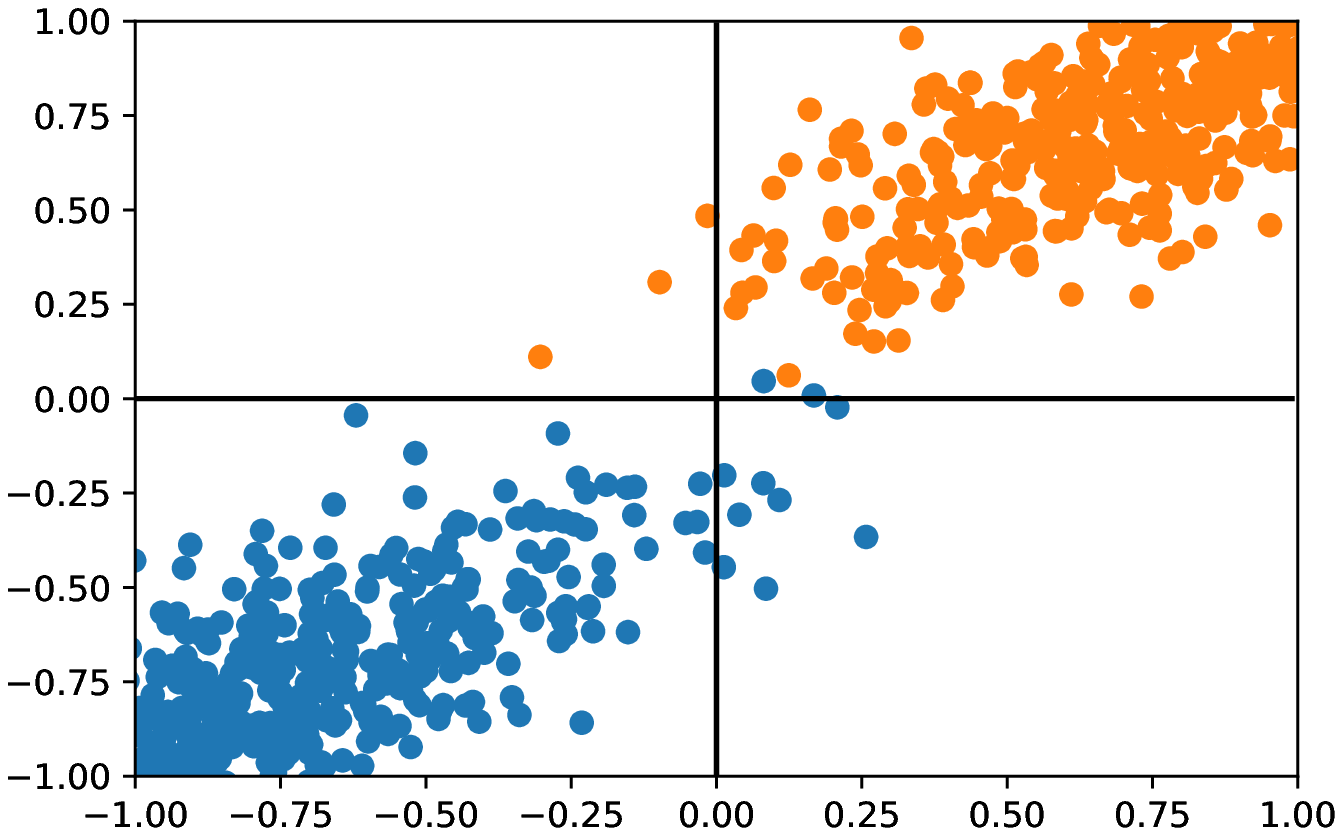}
    \caption{SNRQ ($\alpha\!=\!2$,  $\beta\!=\!50$)}
  \end{subfigure}
    \begin{subfigure}[t]{.3\linewidth}
    \centering\includegraphics[width=4cm]{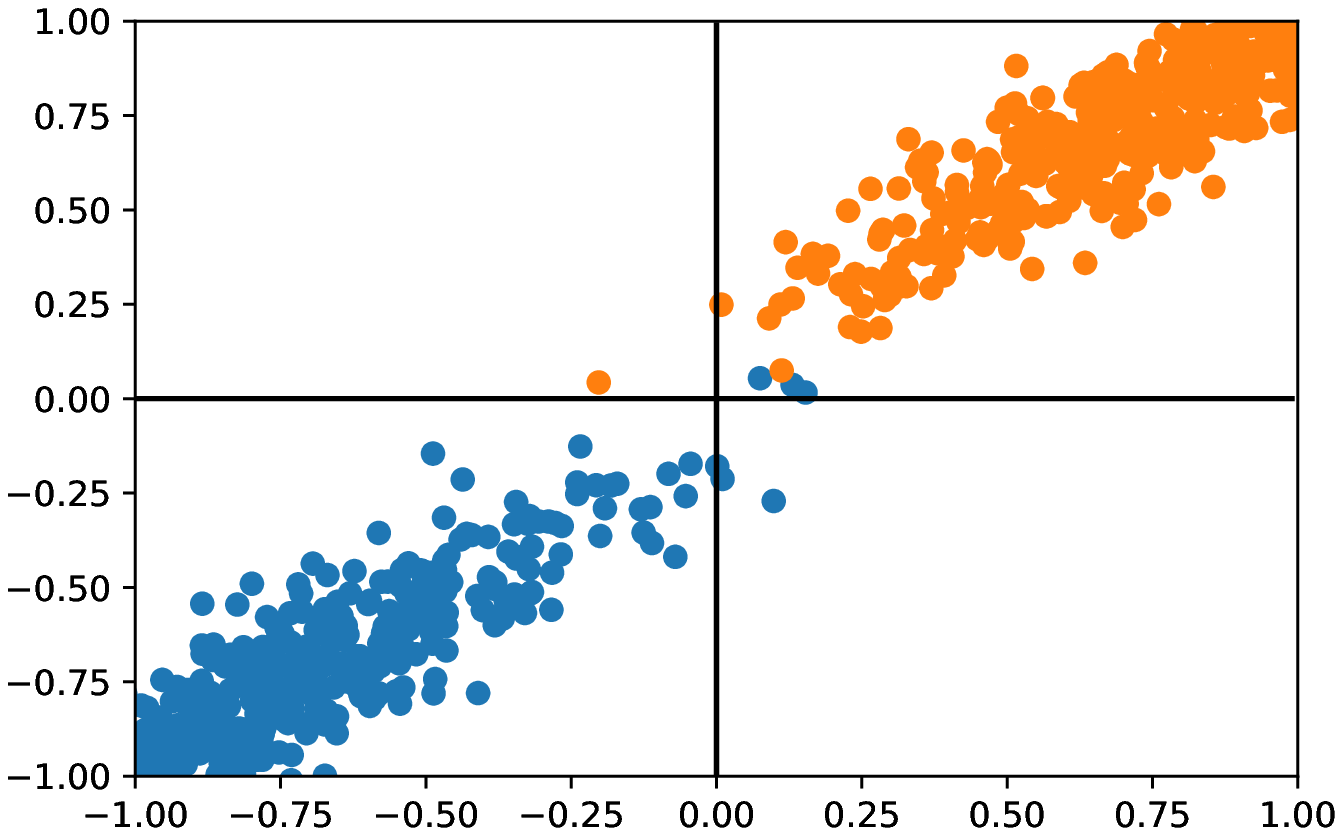}
    \caption{SNRQ ($\alpha\!=\!4$,  $\beta\!=\!50$)}
  \end{subfigure}
  \caption{Visualization of the ITQ \& SNRQ performance in projecting the data to a space with less quantization loss on a 2D toy dataset: (a) The reference 2-D data,  (b) data rotated \& projected by ITQ, (c-f) Data rotated \& projected by SNRQ using  different values of $\beta$ \& $\alpha$. Data points are pushed towards Hamming vertices corrupting the neighbour structure of data.}%
  \label{fig:3}%
\end{figure*}

\textbf{Comparison between NRQ and SNRQ:} Here we compare NRQ and SNRQ both in terms of run-time and performance. If we directly update the projection matrix, the training becomes significantly time consuming. To address this, we proposed sequential update scheme which achieves the same performance while significantly reduces the time complexity. The training times of NRQ and SNRQ for CIFAR-10 dataset  are shown in Fig. \ref{fig:2}. Although  based on Fig. \ref{fig:2}, SNRQ is significantly faster compared with NRQ, as can be seen from Tables \ref{table:table1}, \ref{table:table2}, \ref{table:table3}, and \ref{table:table4} their performance is  similar. The time difference for training may be quite significant for petabyte archives with gigapixel files like satellite imaging and digital pathology where images are commonly quite large, $>$ 50,000 by 50,000 pixels. For the latter each patient often comes with many images.

\textbf{SNRQ vs. ITQ -- }
 To validate how employing a non-rigid  transformation acts compared with rigid one, we generated a synthetic toy dataset. Fig. \ref{fig:3} shows the transformed data by ITQ and SNRQ on this toy dataset under for values of $\alpha$ and $\beta$. Fig. \ref{fig:3} part (a) represents the toy data. Fig. \ref{fig:3} part (b) shows the same data after transformation by ITQ. Figs. \ref{fig:3} part (c), (d), (e), and (f) show the transformed data using SNRQ for different values of $\alpha$ and $\beta$. This graph reveals role of these regularization parameters in SNRQ algorithm which is controlling the trade-off between preserving neighbourhood structure of data and pushing data points. They can cause unsafe quantization. The smaller the $\beta$ (larger the $\alpha$) the unsafer the quantization.  The quantization loss is increasingly reduced  by corrupting neighbourhood.

\textbf{Comparison of SNRQ with deep unsupervised hashing algorithms --} Recently deep learning has been applied to unsupervised hashing including  DH \cite{erin2015deep}, UHBDNN \cite{do2016learning}, DeepBit \cite{lin2016learning}, and SADH \cite{shen2018unsupervised}. To compare SNRQ with Deep unsupervised hashing methods, we employ NUS-WIDE in the setting same as Table \ref{table:table4}. As  Table \ref{table:table5} shows, SNRQ outperforms UHBDNN and DeepBit by a considerable gap. Compared with SADH, results are highly competitive. Considering DeepBit and SADH are feeding images into a CNN and learn the features, which implies that the quality of features used in SADH, and DeepBit are expectedly better than VGG-F extracted features, it is quite impressive that SNRQ is superior or delivering on par results.  
\begin{table}[!h]
\centering
\footnotesize
\caption{SNRQ compared with deep unsupervised hashing algorithms for NUS-WIDE dataset. The terms R+ and V+ mean the respective algorithm works on raw images and vector data (images after feature extraction) respectively. 
}
\begin{tabular}{|l|ccc|ccc|}
\hline
                  & \multicolumn{3}{c|}{mAP \%}    & \multicolumn{3}{c|}{precision \% @5000}                                                                                                                                                                      \\ \hline
Method                    & \multicolumn{1}{c|}{16 bits}       & \multicolumn{1}{c|}{32 bits}       & \multicolumn{1}{c|}{64 bits}      & \multicolumn{1}{c|}{16 bits}  & \multicolumn{1}{c|}{32 bits}       & \multicolumn{1}{c|}{64 bits}             \\ \hline\hline  
V+UHBDN               & 54.26                                  & 51.72                              & 54.74                              &70.18     &69.60        &72.74                                            
\\ \hline
 R+DeepBit                     &39.22                                  & 40.32                              &   42.06                             & 45.54    & 51.34   & 57.72                                                       \\ \hline
 R+SADH                    &60.14                                   &57.99                                &56.33                                 & \textbf{71.45}    & \textbf{73.88}    & \textbf{75.04}                                                 \\ \hline

V+SNRQ             &\textbf{61.78}                            & \textbf{62.74}                              &\textbf{62.60}                           &70.83        &72.33          &73.37               \\ \hline
\end{tabular}
\label{table:table5}
\end{table}


\textbf{Ablation study --} 
We set $\alpha=3$ and $\beta=0.01$ for all experiments and  datasets. Based on the following experiments, although $\beta$ and $\alpha$ regulates the trade-off between preserving the neighbourhood structure and the quality of quantization, SNRQ is robust to changes. Results in  Table \ref{table:table6} show the SNRQ performance on MNIST and LabelMe-12-50k datasets for different values $\beta= 0.01, 0.05, 0.1$, and $0.5$ for the 16 bit setting. Even the worst performance after changing $\beta=0.01$ is still state-of-art.  We also tested the robustness of SNRQ to changes in $\alpha$. We should not use $\alpha=1$ as it removes the $\mathbf{Q}$ from objective. For $\alpha$, first the $\beta$ value is set to 0.01, then we evaluated SNRQ for  $\alpha=$2, 3, 4 and 5 in Table \ref{table:table6}. The performance for $\alpha=$2,3, 4 and 5 is consistently high reaching highest for $\alpha=4$. Considering the proposed objective function in Eq. \ref{eq:3}, if we put $\alpha = 0$, there will be no quantization step, and the problem becomes a PCA-like objective function where the orthogonality constraint has been smoothed. Similar to PCA, this would lead to poor binary representations due to accumulated quantization error. For  $\beta=0$, the projection matrix $\mathbf{W}$ becomes too dominant and destroys the neighbourhood of data leading to loss of information. Besides, from an optimization point of view, for $\alpha = 0$ and $\beta=0$, the objective function would become a quadratic function where, given the fact that covariance matrix is positive semi-definite, the maximization is unbounded. As a result, $\alpha = 0$ and $\beta=0$ do not make sense in the context of this paper.

\textbf{Achieving non-rigid  projections --} We achieved a non-rigid transformation by relaxing orthogonality constraint on projection matrix. We used the well-known soft orthogonality (SO) $\|\mathbf{W}^T \mathbf{W}\!-\!\mathbf{I}_K\|^2_F$. However, there are other ways of achieving this non-rigid projection matrix \cite{bansal2018can}. We also tried double soft orthogonality (DSO) $\|\mathbf{W}^T \mathbf{W}\!-\!\mathbf{I}_K\|^2_F+\|\mathbf{W} \mathbf{W}^T\!-\!\mathbf{I}_D\|^2_F$, and
Mutual Coherence (MC) $\|\mathbf{W}^T \mathbf{W}\!-\!\mathbf{I}_K\|^2_\infty$ where we resorted to auto differentiation for implementation. Based on Table  \ref{table:table7}, the simple soft orthogonality (SO) is achieving better performance in most cases. 
\begin{table}[htb]
\centering
\footnotesize
\caption{Comparison of retrieval performance for MNIST (GIST 512-D) and LabelMe-12-50k (4096-D VGG-FC7) datasets based on mAP for different values of $\beta$ and $\alpha$.}
\begin{tabular}{|l|cccc|}
\hline
     
                  & \multicolumn{4}{c|}{16 bits, $\alpha=3$ }                                                                                                              \\ \hline 
Dataset
                    & \multicolumn{1}{c|}{$\beta=0.01$}       & \multicolumn{1}{c|}{$\beta=0.05$}       & \multicolumn{1}{c|}{$\beta=0.1$}     & \multicolumn{1}{c|}{$\beta=0.5$} 
                   \\ \hline                          
MNIST                       & 64.70                                  & 64.29                         &\textbf{67.17}        
& 66.38                                                                   \\ \hline
LabelMe                       & \textbf{26.08}                                 & 25.29                         & 25.29    
& 25.10                                                                 \\ \hline
 
 \hline
                  & \multicolumn{4}{c|}{16 bits, $\beta=0.01$ }                                                                                                               \\ \hline  
Dataset
                     &\multicolumn{1}{c|}{$\alpha=2$}       & \multicolumn{1}{c|}{$\alpha=3$}     & \multicolumn{1}{c|}{$\alpha=4$} 
                     & \multicolumn{1}{c|}{$\alpha=5$}       \\ \hline                            
MNIST                                                      &63.3                           &  63.18    
&  \textbf{65.58}     &  65.30                                                                \\ \hline
 
LabelMe                                              & 24.20                          &  24.54
&  \textbf{25.73}      &  25.63                                                             \\ \hline

\end{tabular}
\label{table:table6}
\end{table}
\begin{table}[htb]
\centering
\footnotesize
\caption{Comparison of SO, DSO, and MC for obtaining non-rigid transformation on NCT-CRC-HE-100K dataset based on mAP and precision@1000.}
\begin{tabular}{|l|ccc|ccc|}
\hline
                  & \multicolumn{3}{c|}{mAP \%}    & \multicolumn{3}{c|}{precision \% @1000}                                                                                                                                                                      \\ \hline
Method                    & \multicolumn{1}{c|}{16 bits}       & \multicolumn{1}{c|}{32 bits}       & \multicolumn{1}{c|}{64 bits}      & \multicolumn{1}{c|}{16 bits}  & \multicolumn{1}{c|}{32 bits
}       & \multicolumn{1}{c|}{64 bits}             \\ \hline\hline  
SO                        &71.34                                  & \textbf{76.83}                               & 76.57                                 &\textbf{82.94}    &\textbf{86.13  }     &\textbf{86.30}                                        
\\ \hline
 DSO                        & 71.07                                  & 76.29                                 & 75.40                               &81.59    &85.60    &86.16                                                       \\ \hline
MC                      &\textbf{72.43}                                    & 73.94                               & \textbf{76.86 }                   &82.70    &83.72     & 85.86                                                   \\ \hline

\end{tabular}
\label{table:table7}
\end{table}

\section{Conclusions} 
We introduced Sequential Non-rigid Quantization (SNRQ). The backbone of SNRQ is based on ITQ (iterative quantization)  \cite{gong2012iterative}. Although the ideas presented in ITQ are interesting, we argued that learning projection and rotation in two separate steps could be sub-optimal. Furthermore, a rigid transformation may not be enough for reducing quantization to the ultimate limit. Motivated by these limitations, we proposed an algorithm to reduce both dimensionality and quantization loss simultaneously. We also employed a non-rigid transformation to push for quantization beyond rotation. Employing non-rigid transformations is generally against intuition. It does not preserve the neighborhood of data (and all these efforts for reducing quantization error is to preserve more neighborhood after binarization). However, we showed that corrupting neighborhood in favor of reducing quantization eventually leads to better codes. An efficient nested coordinate descent algorithm was employed to update all three matrices. The results on five public datasets totaling almost half a million images showed that the proposed method outperforms the state-of-art linear hashing methods. 

As the future work, we plan to extend the idea of binary representation learning using the rigid and non-rigid transformations to the deep architectures. In this framework, while the non-rigid transformations can be realized using regularization terms similar to the soft orthogonality in Eq. \ref{eq:3}. For the rigid transformation, one can project the gradient on to orthogonal feasible set using the projection operation proposed in \cite{manton2002optimization} to update the rotation matrix. Besides,  it has been recently demonstrated that hashing can be used in the network quantization task \cite{gholami2021survey}. This would be interesting to see how the proposed quantization method performs for network quantization.

\bibliographystyle{model2-names}
\bibliography{refs}

\begin{thebibliography}{38}
\expandafter\ifx\csname natexlab\endcsname\relax\def\natexlab#1{#1}\fi
\providecommand{\url}[1]{\texttt{#1}}
\providecommand{\href}[2]{#2}
\providecommand{\path}[1]{#1}
\providecommand{\DOIprefix}{doi:}
\providecommand{\ArXivprefix}{arXiv:}
\providecommand{\URLprefix}{URL: }
\providecommand{\Pubmedprefix}{pmid:}
\providecommand{\doi}[1]{\href{http://dx.doi.org/#1}{\path{#1}}}
\providecommand{\Pubmed}[1]{\href{pmid:#1}{\path{#1}}}
\providecommand{\bibinfo}[2]{#2}
\ifx\xfnm\relax \def\xfnm[#1]{\unskip,\space#1}\fi
\bibitem[{{Andoni} and {Indyk}(2006)}]{4031381}
\bibinfo{author}{{Andoni}, A.}, \bibinfo{author}{{Indyk}, P.},
  \bibinfo{year}{2006}.
\newblock \bibinfo{title}{Near-optimal hashing algorithms for approximate
  nearest neighbor in high dimensions}, in: \bibinfo{booktitle}{2006 47th
  Annual IEEE Symposium on Foundations of Computer Science (FOCS'06)}, pp.
  \bibinfo{pages}{459--468}.
\newblock \DOIprefix\doi{10.1109/FOCS.2006.49}.
\bibitem[{Bansal et~al.(2018)Bansal, Chen and Wang}]{bansal2018can}
\bibinfo{author}{Bansal, N.}, \bibinfo{author}{Chen, X.},
  \bibinfo{author}{Wang, Z.}, \bibinfo{year}{2018}.
\newblock \bibinfo{title}{Can we gain more from orthogonality regularizations
  in training deep networks?}, in: \bibinfo{booktitle}{Advances in Neural
  Information Processing Systems}, pp. \bibinfo{pages}{4261--4271}.
\bibitem[{Carreira-Perpinan and Raziperchikolaei(2015)}]{carreira2015hashing}
\bibinfo{author}{Carreira-Perpinan, M.A.}, \bibinfo{author}{Raziperchikolaei,
  R.}, \bibinfo{year}{2015}.
\newblock \bibinfo{title}{Hashing with binary autoencoders}, in:
  \bibinfo{booktitle}{Proceedings of the IEEE Conference on Computer Vision and
  Pattern Recognition (CVPR)}.
\bibitem[{Chatfield et~al.(2014)Chatfield, Simonyan, Vedaldi and
  Zisserman}]{BMVC.28.6}
\bibinfo{author}{Chatfield, K.}, \bibinfo{author}{Simonyan, K.},
  \bibinfo{author}{Vedaldi, A.}, \bibinfo{author}{Zisserman, A.},
  \bibinfo{year}{2014}.
\newblock \bibinfo{title}{Return of the devil in the details: Delving deep into
  convolutional nets}, in: \bibinfo{booktitle}{Proceedings of the British
  Machine Vision Conference}, \bibinfo{publisher}{BMVA Press}.
\newblock \DOIprefix\doi{http://dx.doi.org/10.5244/C.28.6}.
\bibitem[{Chu et~al.(2019)Chu, Gong, Chen, Guo, Han and
  Ding}]{chu2019optimized}
\bibinfo{author}{Chu, C.}, \bibinfo{author}{Gong, D.}, \bibinfo{author}{Chen,
  K.}, \bibinfo{author}{Guo, Y.}, \bibinfo{author}{Han, J.},
  \bibinfo{author}{Ding, G.}, \bibinfo{year}{2019}.
\newblock \bibinfo{title}{Optimized projection for hashing}.
\newblock \bibinfo{journal}{Pattern Recognition Letters} \bibinfo{volume}{117},
  \bibinfo{pages}{169--178}.
\bibitem[{Chua et~al.(2009)Chua, Tang, Hong, Li, Luo and
  Zheng}]{10.1145/1646396.1646452}
\bibinfo{author}{Chua, T.S.}, \bibinfo{author}{Tang, J.},
  \bibinfo{author}{Hong, R.}, \bibinfo{author}{Li, H.}, \bibinfo{author}{Luo,
  Z.}, \bibinfo{author}{Zheng, Y.}, \bibinfo{year}{2009}.
\newblock \bibinfo{title}{Nus-wide: a real-world web image database from
  national university of singapore}, in: \bibinfo{booktitle}{Proceedings of the
  ACM international conference on image and video retrieval}, pp.
  \bibinfo{pages}{1--9}.
\bibitem[{Do et~al.(2016)Do, Doan and Cheung}]{do2016learning}
\bibinfo{author}{Do, T.T.}, \bibinfo{author}{Doan, A.D.},
  \bibinfo{author}{Cheung, N.M.}, \bibinfo{year}{2016}.
\newblock \bibinfo{title}{Learning to hash with binary deep neural network},
  in: \bibinfo{booktitle}{European Conference on Computer Vision},
  \bibinfo{organization}{Springer}. pp. \bibinfo{pages}{219--234}.
\bibitem[{Gholami et~al.(2021)Gholami, Kim, Dong, Yao, Mahoney and
  Keutzer}]{gholami2021survey}
\bibinfo{author}{Gholami, A.}, \bibinfo{author}{Kim, S.},
  \bibinfo{author}{Dong, Z.}, \bibinfo{author}{Yao, Z.},
  \bibinfo{author}{Mahoney, M.W.}, \bibinfo{author}{Keutzer, K.},
  \bibinfo{year}{2021}.
\newblock \bibinfo{title}{A survey of quantization methods for efficient neural
  network inference}.
\newblock \bibinfo{journal}{arXiv preprint arXiv:2103.13630} .
\bibitem[{Gong et~al.(2012)Gong, Kumar, Verma and Lazebnik}]{gong2012angular}
\bibinfo{author}{Gong, Y.}, \bibinfo{author}{Kumar, S.},
  \bibinfo{author}{Verma, V.}, \bibinfo{author}{Lazebnik, S.},
  \bibinfo{year}{2012}.
\newblock \bibinfo{title}{Angular quantization-based binary codes for fast
  similarity search}, in: \bibinfo{booktitle}{Advances in neural information
  processing systems}, pp. \bibinfo{pages}{1196--1204}.
\bibitem[{Gong et~al.(2013)Gong, Lazebnik, Gordo and
  Perronnin}]{gong2012iterative}
\bibinfo{author}{Gong, Y.}, \bibinfo{author}{Lazebnik, S.},
  \bibinfo{author}{Gordo, A.}, \bibinfo{author}{Perronnin, F.},
  \bibinfo{year}{2013}.
\newblock \bibinfo{title}{Iterative quantization: A procrustean approach to
  learning binary codes for large-scale image retrieval}.
\newblock \bibinfo{journal}{IEEE Transactions on Pattern Analysis and Machine
  Intelligence} \bibinfo{volume}{35}, \bibinfo{pages}{2916--2929}.
\newblock \DOIprefix\doi{10.1109/TPAMI.2012.193}.
\bibitem[{He et~al.(2013)He, Wen and Sun}]{he2013k}
\bibinfo{author}{He, K.}, \bibinfo{author}{Wen, F.}, \bibinfo{author}{Sun, J.},
  \bibinfo{year}{2013}.
\newblock \bibinfo{title}{K-means hashing: An affinity-preserving quantization
  method for learning binary compact codes}, in:
  \bibinfo{booktitle}{Proceedings of the IEEE Conference on Computer Vision and
  Pattern Recognition (CVPR)}.
\bibitem[{He et~al.(2019)He, Wang and Cheng}]{He_2019_CVPR}
\bibinfo{author}{He, X.}, \bibinfo{author}{Wang, P.}, \bibinfo{author}{Cheng,
  J.}, \bibinfo{year}{2019}.
\newblock \bibinfo{title}{K-nearest neighbors hashing}, in:
  \bibinfo{booktitle}{The IEEE Conference on Computer Vision and Pattern
  Recognition (CVPR)}.
\bibitem[{Hemati et~al.(2020)Hemati, Mehdizavareh, Chenouri and
  Tizhoosh}]{hemati2020non}
\bibinfo{author}{Hemati, S.}, \bibinfo{author}{Mehdizavareh, M.H.},
  \bibinfo{author}{Chenouri, S.}, \bibinfo{author}{Tizhoosh, H.R.},
  \bibinfo{year}{2020}.
\newblock \bibinfo{title}{A non-alternating graph hashing algorithm for large
  scale image search}.
\newblock \bibinfo{journal}{arXiv preprint arXiv:2012.13138} .
\bibitem[{Heo et~al.(2012)Heo, Lee, He, Chang and Yoon}]{heo2012spherical}
\bibinfo{author}{Heo, J.P.}, \bibinfo{author}{Lee, Y.}, \bibinfo{author}{He,
  J.}, \bibinfo{author}{Chang, S.F.}, \bibinfo{author}{Yoon, S.E.},
  \bibinfo{year}{2012}.
\newblock \bibinfo{title}{Spherical hashing}, in: \bibinfo{booktitle}{2012 IEEE
  Conference on Computer Vision and Pattern Recognition}, pp.
  \bibinfo{pages}{2957--2964}.
\newblock \DOIprefix\doi{10.1109/CVPR.2012.6248024}.
\bibitem[{Hoang et~al.(2020)Hoang, Do, Le, Le-Tan and Cheung}]{HOANG2020102852}
\bibinfo{author}{Hoang, T.}, \bibinfo{author}{Do, T.T.}, \bibinfo{author}{Le,
  H.}, \bibinfo{author}{Le-Tan, D.K.}, \bibinfo{author}{Cheung, N.M.},
  \bibinfo{year}{2020}.
\newblock \bibinfo{title}{Simultaneous compression and quantization: A joint
  approach for efficient unsupervised hashing}.
\newblock \bibinfo{journal}{Computer Vision and Image Understanding}
  \bibinfo{volume}{191}, \bibinfo{pages}{102852}.
\bibitem[{Hu et~al.(2018)Hu, Nie and Li}]{hu2018discrete}
\bibinfo{author}{Hu, D.}, \bibinfo{author}{Nie, F.}, \bibinfo{author}{Li, X.},
  \bibinfo{year}{2018}.
\newblock \bibinfo{title}{Discrete spectral hashing for efficient similarity
  retrieval}.
\newblock \bibinfo{journal}{IEEE Transactions on Image Processing}
  \bibinfo{volume}{28}, \bibinfo{pages}{1080--1091}.
\bibitem[{Kong and Li(2012)}]{kong2012isotropic}
\bibinfo{author}{Kong, W.}, \bibinfo{author}{Li, W.j.}, \bibinfo{year}{2012}.
\newblock \bibinfo{title}{Isotropic hashing}, in: \bibinfo{editor}{Pereira,
  F.}, \bibinfo{editor}{Burges, C.J.C.}, \bibinfo{editor}{Bottou, L.},
  \bibinfo{editor}{Weinberger, K.Q.} (Eds.), \bibinfo{booktitle}{Advances in
  Neural Information Processing Systems}, \bibinfo{publisher}{Curran
  Associates, Inc.}
\bibitem[{Krizhevsky(2009)}]{krizhevsky2009learning}
\bibinfo{author}{Krizhevsky, A.}, \bibinfo{year}{2009}.
\newblock \bibinfo{title}{Learning multiple layers of features from tiny
  images} , \bibinfo{pages}{32--33}.
\bibitem[{LeCun et~al.(1998)LeCun, Bottou, Bengio and
  Haffner}]{lecun1998gradient}
\bibinfo{author}{LeCun, Y.}, \bibinfo{author}{Bottou, L.},
  \bibinfo{author}{Bengio, Y.}, \bibinfo{author}{Haffner, P.},
  \bibinfo{year}{1998}.
\newblock \bibinfo{title}{Gradient-based learning applied to document
  recognition}.
\newblock \bibinfo{journal}{Proceedings of the IEEE} \bibinfo{volume}{86},
  \bibinfo{pages}{2278--2324}.
\bibitem[{Li et~al.(2017)Li, Hu and Nie}]{li2017large}
\bibinfo{author}{Li, X.}, \bibinfo{author}{Hu, D.}, \bibinfo{author}{Nie, F.},
  \bibinfo{year}{2017}.
\newblock \bibinfo{title}{Large graph hashing with spectral rotation}, in:
  \bibinfo{booktitle}{Thirty-First AAAI Conference on Artificial Intelligence}.
\bibitem[{Lin et~al.(2016)Lin, Lu, Chen and Zhou}]{lin2016learning}
\bibinfo{author}{Lin, K.}, \bibinfo{author}{Lu, J.}, \bibinfo{author}{Chen,
  C.S.}, \bibinfo{author}{Zhou, J.}, \bibinfo{year}{2016}.
\newblock \bibinfo{title}{Learning compact binary descriptors with unsupervised
  deep neural networks}, in: \bibinfo{booktitle}{Proceedings of the IEEE
  Conference on Computer Vision and Pattern Recognition}, pp.
  \bibinfo{pages}{1183--1192}.
\bibitem[{Liong et~al.(2015)Liong, Lu, Wang, Moulin and Zhou}]{erin2015deep}
\bibinfo{author}{Liong, V.E.}, \bibinfo{author}{Lu, J.}, \bibinfo{author}{Wang,
  G.}, \bibinfo{author}{Moulin, P.}, \bibinfo{author}{Zhou, J.},
  \bibinfo{year}{2015}.
\newblock \bibinfo{title}{Deep hashing for compact binary codes learning}, in:
  \bibinfo{booktitle}{2015 IEEE Conference on Computer Vision and Pattern
  Recognition (CVPR)}, pp. \bibinfo{pages}{2475--2483}.
\newblock \DOIprefix\doi{10.1109/CVPR.2015.7298862}.
\bibitem[{Liu et~al.(2014)Liu, Mu, Kumar and Chang}]{liu2014discrete}
\bibinfo{author}{Liu, W.}, \bibinfo{author}{Mu, C.}, \bibinfo{author}{Kumar,
  S.}, \bibinfo{author}{Chang, S.F.}, \bibinfo{year}{2014}.
\newblock \bibinfo{title}{Discrete graph hashing}, in:
  \bibinfo{editor}{Ghahramani, Z.}, \bibinfo{editor}{Welling, M.},
  \bibinfo{editor}{Cortes, C.}, \bibinfo{editor}{Lawrence, N.},
  \bibinfo{editor}{Weinberger, K.Q.} (Eds.), \bibinfo{booktitle}{Advances in
  Neural Information Processing Systems}, \bibinfo{publisher}{Curran
  Associates, Inc.}
\bibitem[{Liu et~al.(2011)Liu, Wang, Kumar and Chang}]{liu2011hashing}
\bibinfo{author}{Liu, W.}, \bibinfo{author}{Wang, J.}, \bibinfo{author}{Kumar,
  S.}, \bibinfo{author}{Chang, S.F.}, \bibinfo{year}{2011}.
\newblock \bibinfo{title}{Hashing with graphs}, in: \bibinfo{booktitle}{Icml}.
\bibitem[{Macenko et~al.(2009)Macenko, Niethammer, Marron, Borland, Woosley,
  Guan, Schmitt and Thomas}]{macenko2009method}
\bibinfo{author}{Macenko, M.}, \bibinfo{author}{Niethammer, M.},
  \bibinfo{author}{Marron, J.S.}, \bibinfo{author}{Borland, D.},
  \bibinfo{author}{Woosley, J.T.}, \bibinfo{author}{Guan, X.},
  \bibinfo{author}{Schmitt, C.}, \bibinfo{author}{Thomas, N.E.},
  \bibinfo{year}{2009}.
\newblock \bibinfo{title}{A method for normalizing histology slides for
  quantitative analysis}, in: \bibinfo{booktitle}{2009 IEEE International
  Symposium on Biomedical Imaging: From Nano to Macro},
  \bibinfo{organization}{IEEE}. pp. \bibinfo{pages}{1107--1110}.
\bibitem[{Manton(2002)}]{manton2002optimization}
\bibinfo{author}{Manton, J.H.}, \bibinfo{year}{2002}.
\newblock \bibinfo{title}{Optimization algorithms exploiting unitary
  constraints}.
\newblock \bibinfo{journal}{IEEE Transactions on Signal Processing}
  \bibinfo{volume}{50}, \bibinfo{pages}{635--650}.
\bibitem[{Paulev{\'e} et~al.(2010)Paulev{\'e}, J{\'e}gou and
  Amsaleg}]{pauleve2010locality}
\bibinfo{author}{Paulev{\'e}, L.}, \bibinfo{author}{J{\'e}gou, H.},
  \bibinfo{author}{Amsaleg, L.}, \bibinfo{year}{2010}.
\newblock \bibinfo{title}{{Locality sensitive hashing: a comparison of hash
  function types and querying mechanisms}}.
\newblock \bibinfo{journal}{{Pattern Recognition Letters}}
  \bibinfo{volume}{31}, \bibinfo{pages}{1348--1358}.
\bibitem[{Sch{\"o}nemann(1966)}]{schonemann1966generalized}
\bibinfo{author}{Sch{\"o}nemann, P.H.}, \bibinfo{year}{1966}.
\newblock \bibinfo{title}{A generalized solution of the orthogonal procrustes
  problem}.
\newblock \bibinfo{journal}{Psychometrika} \bibinfo{volume}{31},
  \bibinfo{pages}{1--10}.
\bibitem[{Shao et~al.(2012)Shao, Wu, Ouyang and Zhang}]{shao2012sparse}
\bibinfo{author}{Shao, J.}, \bibinfo{author}{Wu, F.}, \bibinfo{author}{Ouyang,
  C.}, \bibinfo{author}{Zhang, X.}, \bibinfo{year}{2012}.
\newblock \bibinfo{title}{Sparse spectral hashing}.
\newblock \bibinfo{journal}{Pattern Recognition Letters} \bibinfo{volume}{33},
  \bibinfo{pages}{271--277}.
\newblock \DOIprefix\doi{https://doi.org/10.1016/j.patrec.2011.10.018}.
\bibitem[{Shen et~al.(2018)Shen, Xu, Liu, Yang, Huang and
  Shen}]{shen2018unsupervised}
\bibinfo{author}{Shen, F.}, \bibinfo{author}{Xu, Y.}, \bibinfo{author}{Liu,
  L.}, \bibinfo{author}{Yang, Y.}, \bibinfo{author}{Huang, Z.},
  \bibinfo{author}{Shen, H.T.}, \bibinfo{year}{2018}.
\newblock \bibinfo{title}{Unsupervised deep hashing with similarity-adaptive
  and discrete optimization}.
\newblock \bibinfo{journal}{IEEE Transactions on Pattern Analysis and Machine
  Intelligence} \bibinfo{volume}{40}, \bibinfo{pages}{3034--3044}.
\newblock \DOIprefix\doi{10.1109/TPAMI.2018.2789887}.
\bibitem[{Simonyan and Zisserman(2014)}]{simonyan2014very}
\bibinfo{author}{Simonyan, K.}, \bibinfo{author}{Zisserman, A.},
  \bibinfo{year}{2014}.
\newblock \bibinfo{title}{Very deep convolutional networks for large-scale
  image recognition}.
\newblock \bibinfo{journal}{arXiv preprint arXiv:1409.1556} .
\bibitem[{Tan and Le(2019)}]{tan2019efficientnet}
\bibinfo{author}{Tan, M.}, \bibinfo{author}{Le, Q.V.}, \bibinfo{year}{2019}.
\newblock \bibinfo{title}{Efficientnet: Rethinking model scaling for
  convolutional neural networks}.
\newblock \bibinfo{journal}{arXiv preprint arXiv:1905.11946} .
\bibitem[{Uetz and Behnke(2009)}]{uetz2009large}
\bibinfo{author}{Uetz, R.}, \bibinfo{author}{Behnke, S.}, \bibinfo{year}{2009}.
\newblock \bibinfo{title}{Large-scale object recognition with cuda-accelerated
  hierarchical neural networks}, in: \bibinfo{booktitle}{2009 IEEE
  international conference on intelligent computing and intelligent systems},
  \bibinfo{organization}{IEEE}. pp. \bibinfo{pages}{536--541}.
\bibitem[{Wang et~al.(2018)Wang, Zhang, song, Sebe and Shen}]{wang2017survey}
\bibinfo{author}{Wang, J.}, \bibinfo{author}{Zhang, T.}, \bibinfo{author}{song,
  j.}, \bibinfo{author}{Sebe, N.}, \bibinfo{author}{Shen, H.T.},
  \bibinfo{year}{2018}.
\newblock \bibinfo{title}{A survey on learning to hash}.
\newblock \bibinfo{journal}{IEEE Transactions on Pattern Analysis and Machine
  Intelligence} \bibinfo{volume}{40}, \bibinfo{pages}{769--790}.
\newblock \DOIprefix\doi{10.1109/TPAMI.2017.2699960}.
\bibitem[{Weiss et~al.(2009)Weiss, Torralba and Fergus}]{weiss2009spectral}
\bibinfo{author}{Weiss, Y.}, \bibinfo{author}{Torralba, A.},
  \bibinfo{author}{Fergus, R.}, \bibinfo{year}{2009}.
\newblock \bibinfo{title}{Spectral hashing}, in: \bibinfo{editor}{Koller, D.},
  \bibinfo{editor}{Schuurmans, D.}, \bibinfo{editor}{Bengio, Y.},
  \bibinfo{editor}{Bottou, L.} (Eds.), \bibinfo{booktitle}{Advances in Neural
  Information Processing Systems}, \bibinfo{publisher}{Curran Associates, Inc.}
\bibitem[{{Yunchao Gong} et~al.(2015){Yunchao Gong}, {Pawlowski}, {Fei Yang},
  {Brandy}, {Boundev} and {Fergus}}]{7298596}
\bibinfo{author}{{Yunchao Gong}}, \bibinfo{author}{{Pawlowski}, M.},
  \bibinfo{author}{{Fei Yang}}, \bibinfo{author}{{Brandy}, L.},
  \bibinfo{author}{{Boundev}, L.}, \bibinfo{author}{{Fergus}, R.},
  \bibinfo{year}{2015}.
\newblock \bibinfo{title}{Web scale photo hash clustering on a single machine},
  in: \bibinfo{booktitle}{2015 IEEE Conference on Computer Vision and Pattern
  Recognition (CVPR)}, pp. \bibinfo{pages}{19--27}.
\bibitem[{Zhang et~al.(2015)Zhang, Lin, Zhang, Zuo and Zhang}]{zhang2015bit}
\bibinfo{author}{Zhang, R.}, \bibinfo{author}{Lin, L.}, \bibinfo{author}{Zhang,
  R.}, \bibinfo{author}{Zuo, W.}, \bibinfo{author}{Zhang, L.},
  \bibinfo{year}{2015}.
\newblock \bibinfo{title}{Bit-scalable deep hashing with regularized similarity
  learning for image retrieval and person re-identification}.
\newblock \bibinfo{journal}{IEEE Transactions on Image Processing}
  \bibinfo{volume}{24}, \bibinfo{pages}{4766--4779}.
\bibitem[{Zhu et~al.(1997)Zhu, Byrd, Lu and Nocedal}]{zhu1997algorithm}
\bibinfo{author}{Zhu, C.}, \bibinfo{author}{Byrd, R.H.}, \bibinfo{author}{Lu,
  P.}, \bibinfo{author}{Nocedal, J.}, \bibinfo{year}{1997}.
\newblock \bibinfo{title}{Algorithm 778: L-bfgs-b: Fortran subroutines for
  large-scale bound-constrained optimization}.
\newblock \bibinfo{journal}{ACM Transactions on Mathematical Software (TOMS)}
  \bibinfo{volume}{23}, \bibinfo{pages}{550--560}.

\end{thebibliography}

\end{document}